\documentclass[10pt,twocolumn,letterpaper]{article}

\usepackage{cvpr}
\usepackage{times}
\usepackage{epsfig}
\usepackage{graphicx}
\usepackage{amsmath}
\usepackage{amssymb}
\usepackage{enumitem}

\def\ie{\textit{i.e.}~}

\def\etal{\textit{et al.}~}

\usepackage[pagebackref=true,breaklinks=true,letterpaper=true,colorlinks,bookmarks=false,urlcolor=red]{hyperref}
\usepackage{gensymb,graphics,subfigure,inputenc}
\usepackage[linesnumbered,algo2e,boxed]{algorithm2e}
\graphicspath{{Figure/}}
\usepackage[figuresright]{rotating}
\usepackage{amsmath,amssymb,textcomp,subfigure,multirow,upgreek}
\usepackage{booktabs}
\usepackage{color,mdwlist}

\newcommand{\tht}[2]{\begin{tabular}{@{}#1@{}}#2\end{tabular}}
\newcommand{\red}[1]{\textcolor{blue}{#1}}

\usepackage{graphicx,fontawesome}
\def\ci#1{\textcircled{\resizebox{.4em}{!}{#1}}}

\usepackage{caption}
\usepackage{enumitem}
\setitemize{noitemsep,topsep=0pt,parsep=0pt,partopsep=0pt}

\newcommand*\samethanks[1][\value{footnote}]{\footnotemark[#1]}

\cvprfinalcopy % *** Uncomment this line for the final submission

\def\cvprPaperID{3069} % *** Enter the CVPR Paper ID here

\ifcvprfinal\fi
\begin{document}

%%%%%%%%% TITLE
\title{Multi-Channel Attention Selection GAN with Cascaded Semantic Guidance \\ for Cross-View Image Translation}
\author{Hao Tang$^{1,2}$\thanks{Equal contribution.} \ \ \ Dan Xu$^{3}$\samethanks \ \ \ \ Nicu Sebe$^{1,4}$ \ Yanzhi Wang$^5$ \ Jason J. Corso$^6$ \ Yan Yan$^2$\vspace{6pt}\\
$^1$DISI, University of Trento, Trento, Italy \quad
$^2$Texas State University, San Marcos, USA\\
$^3$University of Oxford, Oxford, UK \quad
$^4$Huawei Technologies Ireland, Dublin, Ireland\\
$^5$Northeastern University, Boston, USA \quad
$^6$University of Michigan, Ann Arbor, USA \\
}

\maketitle

\begin{abstract}
	Cross-view image translation is challenging because it involves  images with drastically different views and severe deformation.
	In this paper, we propose a novel approach named Multi-Channel Attention SelectionGAN (SelectionGAN) that makes it possible to generate images of natural scenes in arbitrary viewpoints, based on an image of the scene and a novel semantic map. The proposed SelectionGAN explicitly utilizes the semantic information and consists of two stages. In the first stage, the condition image and the target semantic map are fed into a cycled semantic-guided generation network to produce initial coarse results. In the second stage, we refine the initial results by using a multi-channel attention selection mechanism. Moreover, uncertainty maps automatically learned from attentions are used to guide the pixel loss for better network optimization. Extensive experiments on Dayton~\cite{vo2016localizing}, CVUSA~\cite{workman2015wide} and Ego2Top~\cite{ardeshir2016ego2top} datasets show that our model is able to generate significantly better results than the state-of-the-art methods. The source code, data and trained models are available at~\url{https://github.com/Ha0Tang/SelectionGAN}.
\end{abstract}

\section{Introduction}
\vspace{-0.08cm}
Cross-view image translation is a task that aims at synthesizing new images from one viewpoint to another.  It has been gaining a lot interest especially from computer vision and virtual reality communities, and has been widely investigated in recent years~\cite{tatarchenko2016multi,kim2017learning,zhu2018generative,regmi2018cross,zhai2017predicting,huang2017beyond,park2017transformation,zhou2016view,yang2015weakly}. Earlier works studied this problem using encoder-decoder Convolutional Neural Networks (CNNs) by involving viewpoint codes in the bottle-neck representations for city scene synthesis~\cite{zhou2016view} and 3D object translation~\cite{yang2015weakly}. There also exist some works exploring Generative Adversarial Networks (GAN) for similar tasks~\cite{park2017transformation}. However, these existing works consider an application scenario in which the objects and the scenes have a large degree of overlapping in appearances and views. 
\begin{figure}[!t]\small
	\centering
	\includegraphics[width=0.98\linewidth]{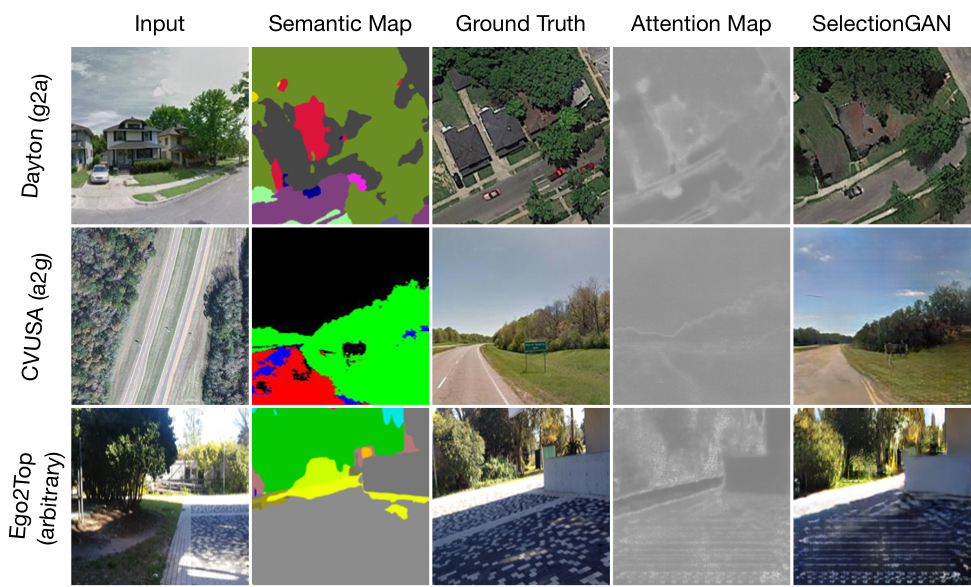}
	\caption{Examples of our cross-view translation results on two public benchmarks \ie Dayton~\cite{vo2016localizing} and CVUSA~\cite{workman2015wide}, and on our self-created large-scale benchmark based on Ego2Top~\cite{ardeshir2016ego2top}.}
	\label{fig:motivation fig.}
	\vspace{-0.05cm}
\end{figure}

\begin{figure*}
	\centering
	\includegraphics[width=0.95\linewidth]{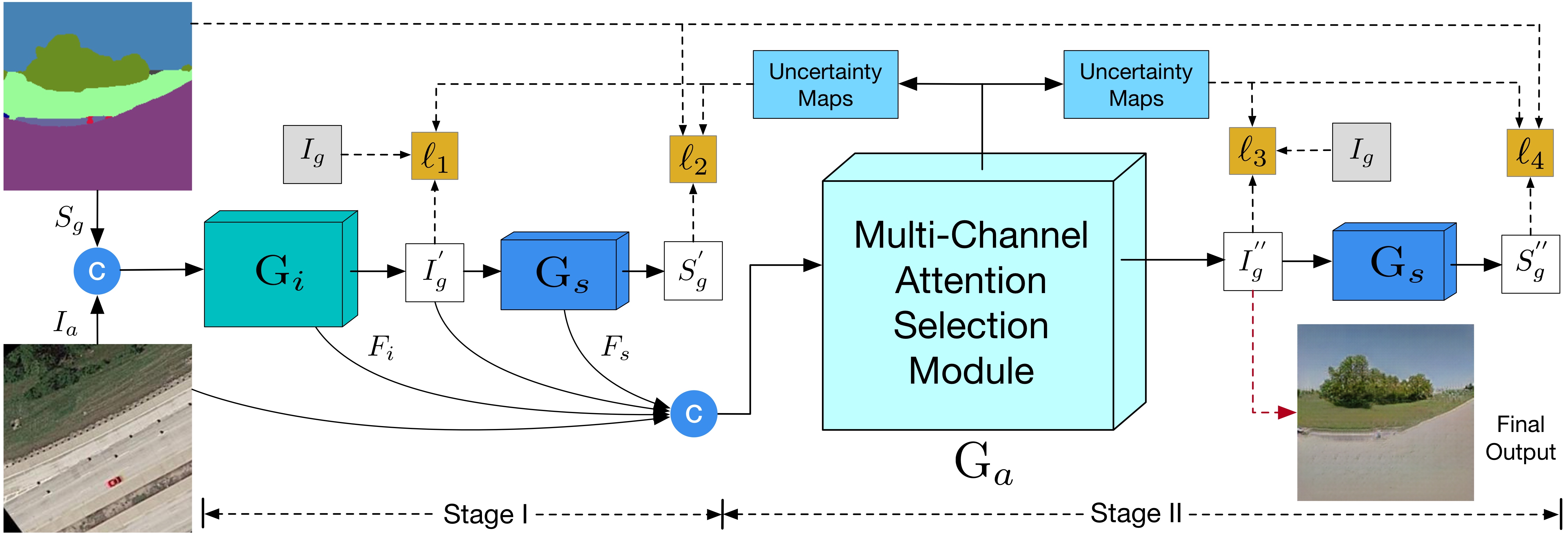}
	\caption{Overview of the proposed SelectionGAN. Stage I presents a cycled semantic-guided generation sub-network which accepts images from one view and conditional semantic maps and simultaneously synthesizes images and semantic maps in another view. Stage II takes the coarse predictions and the learned deep semantic features from stage I, and performs a fine-grained generation using the proposed multi-channel attention selection module.}
	\label{fig:framework}
	\vspace{-0.6cm}
\end{figure*}
\vspace{-0.18cm}
Different from previous works, in this paper, we focus on a more challenging setting in which fields of views have little or even no overlap, leading to significantly distinct structures and appearance distributions for the input source and the output target views, as illustrated in Fig.~\ref{fig:motivation fig.}. To tackle this challenging problem, Regmi and Borji~\cite{regmi2018cross} recently proposed a conditional GAN model which jointly learns the generation in both the image domain and the corresponding semantic domain, and the semantic predictions are further utilized to supervise the image generation. Although this approach performed an interesting exploration, we observe unsatisfactory aspects mainly in the generated scene structure and details, which are due to different reasons. First, since it is always costly to obtain manually annotated semantic labels, the label maps are usually produced from pretrained semantic models from other large-scale segmentation datasets, leading to insufficiently accurate predictions for all the pixels, and thus misguiding the image generation. Second, we argue that the translation with a single phase generation network is not able to capture the complex scene structural relationships between the two views. Third, a three-channel generation space may not be suitable enough for learning a good mapping for this complex synthesis problem. 
Given these problems, could we enlarge the generation space and learn an automatic selection mechanism to synthesize more fine-grained generation results?

Based on these observations, in this paper, we propose a novel Multi-Channel Attention Selection Generative Adversarial Network (SelectionGAN), which contains two generation stages. 
The overall framework of the proposed SelectionGAN is shown in Fig.~\ref{fig:framework}.
In this first stage, we learn a cycled image-semantic generation sub-network, which accepts a pair consisting of an image and the target semantic map, and generates images for the other view, which further fed into a semantic generation network to reconstruct the input semantic maps. This cycled generation adds more strong supervision between the image and semantic domains, facilitating the optimization of the network. 

The coarse outputs from the first generation network, including the input
image, together with the deep feature maps from the
last layer, are input into the second stage networks. Several
intermediate outputs are produced, and simultaneously we
learn a set of multi-channel attention maps with the same
number as the intermediate generations. These attention
maps are used to spatially select from the intermediate generations,
and are combined to synthesize a final output.
Finally, to overcome the inaccurate semantic label issue, the multi-channel attention maps are further used to generate uncertainty maps to guide the reconstruction loss. Through extensive experimental evaluations, we demonstrate that SelectionGAN produces remarkably better results than the baselines such as Pix2pix~\cite{isola2017image}, Zhai \etal \cite{zhai2017predicting}, X-Fork \cite{regmi2018cross} and X-Seq~\cite{regmi2018cross}. Moreover, we establish state-of-the-art results on three different datasets for the arbitrary cross-view image synthesis task. 

Overall, the contributions of this paper are as follows:
\begin{itemize}[leftmargin=*]
	\item A novel multi-channel attention selection GAN framework (SelectionGAN) for the cross-view image translation task is presented. It explores cascaded semantic guidance with a coarse-to-fine inference, and aims at producing a more detailed synthesis from richer and more diverse multiple intermediate generations.
	\item A novel multi-channel attention selection module is proposed, which is utilized to attentively select interested intermediate generations and is able to significantly boost the quality of the final output. The multi-channel attention module also effectively learns uncertainty maps to guide the pixel loss for more robust optimization.
	\item Extensive experiments clearly demonstrate the effectiveness of the proposed SelectionGAN, and show state-of-the-art results on two public benchmarks, \ie Dayton~\cite{vo2016localizing} and CVUSA~\cite{workman2015wide}. Meanwhile, we also create a larger-scale cross-view synthesis benchmark using the data from Ego2Top~\cite{ardeshir2016ego2top}, and present results of multiple baseline models for the research community.
\end{itemize}

\section{Related Work}
\vspace{-0.2cm}

\noindent\textbf{Generative Adversarial Networks (GANs)}
\cite{goodfellow2014generative} have shown the capability of generating better high-quality images \cite{wang2016generative,karras2017progressive,gulrajani2017improved,siarohin2018whitening}, compared to existing methods such as Restricted Boltzmann Machines \cite{hinton2006fast} and Deep Boltzmann Machines \cite{hinton2012better}.
A vanilla GAN model \cite{goodfellow2014generative} has two important components, \ie a generator $G$ and a discriminator $D$.
The goal of $G$ is to generate photo-realistic images from a noise vector, while $D$ is trying to distinguish between a real image and the image generated by $G$.
Although it is successfully used in generating images of high visual fidelity \cite{karras2017progressive,zhang2018self,radford2015unsupervised}, there are still some challenges, \ie how to generate images in a controlled setting. 
To generate domain-specific images, Conditional GAN (CGAN)~\cite{mirza2014conditional} has been proposed.
CGAN usually combines a vanilla GAN and some external information, such as class labels or tags \cite{odena2016semi,odena2016conditional,choi2017stargan,tang2019dual,tang2019attribute}, text descriptions \cite{reed2016learning,han2017stackgan}, human pose \cite{dong2018soft,tang2018gesturegan,neverova2018dense,lakhal2018pose,siarohin2018animating} and reference images \cite{mathieu2015deep,isola2017image}.

\noindent\textbf{Image-to-Image Translation} frameworks adopt input-output data to learn a parametric mapping between inputs and outputs.
For example, Isola \etal \cite{isola2017image} propose Pix2pix, which is a supervised model and uses a CGAN to learn a translation function from input to output image domains.
Zhu~\etal~\cite{zhu2017unpaired} introduce CycleGAN, which targets unpaired image translation using the cycle-consistency loss.
To further improve the generation performance, the attention mechanism has been recently investigated in image translation, such as \cite{chen2018attention,xu2018attngan,tang2019attention,ma2018gan,mejjati2018unsupervised}. However, to the best of our knowledge, our model is the first attempt to incorporate a multi-channel attention selection module within a GAN framework for image-to-image translation task.

\noindent\textbf{Learning Viewpoint Transformations.}
Most existing works on viewpoint transformation have been conducted to synthesize novel views of the same object, such as cars, chairs and tables \cite{dosovitskiy2017learning,tatarchenko2016multi,choy20163d}.
Another group of works explore the cross-view scene image generation, such as \cite{yin2018novel,zhou2016view}.
However, these works focus on the scenario in which the objects and the scenes have a large degree of overlapping in both appearances and views.
Recently, several works started investigating image translation problems with drastically different views and generating a novel scene from a given arbitrary one. This is a more challenging task since different views have little or no overlap. To tackle this problem, Zhai \etal \cite{zhai2017predicting} try to generate panoramic ground-level images from aerial images of the same location by using a convolutional neural network. 
Krishna and Ali~\cite{regmi2018cross} propose a X-Fork and a X-Seq GAN-based structure to address the aerial to street view image translation task using an extra semantic segmentation map. However, these methods are not able to generate satisfactory results due to the drastic difference between source and target views and their model design. To overcome these issues, we aim at a more effective network design, and propose a novel multi-channel attention selection GAN, which allows to automatically select from multiple diverse and rich intermediate generations and thus significantly improves the generation quality.

\begin{figure*}[!t]
	\centering
	\includegraphics[width=0.95\linewidth]{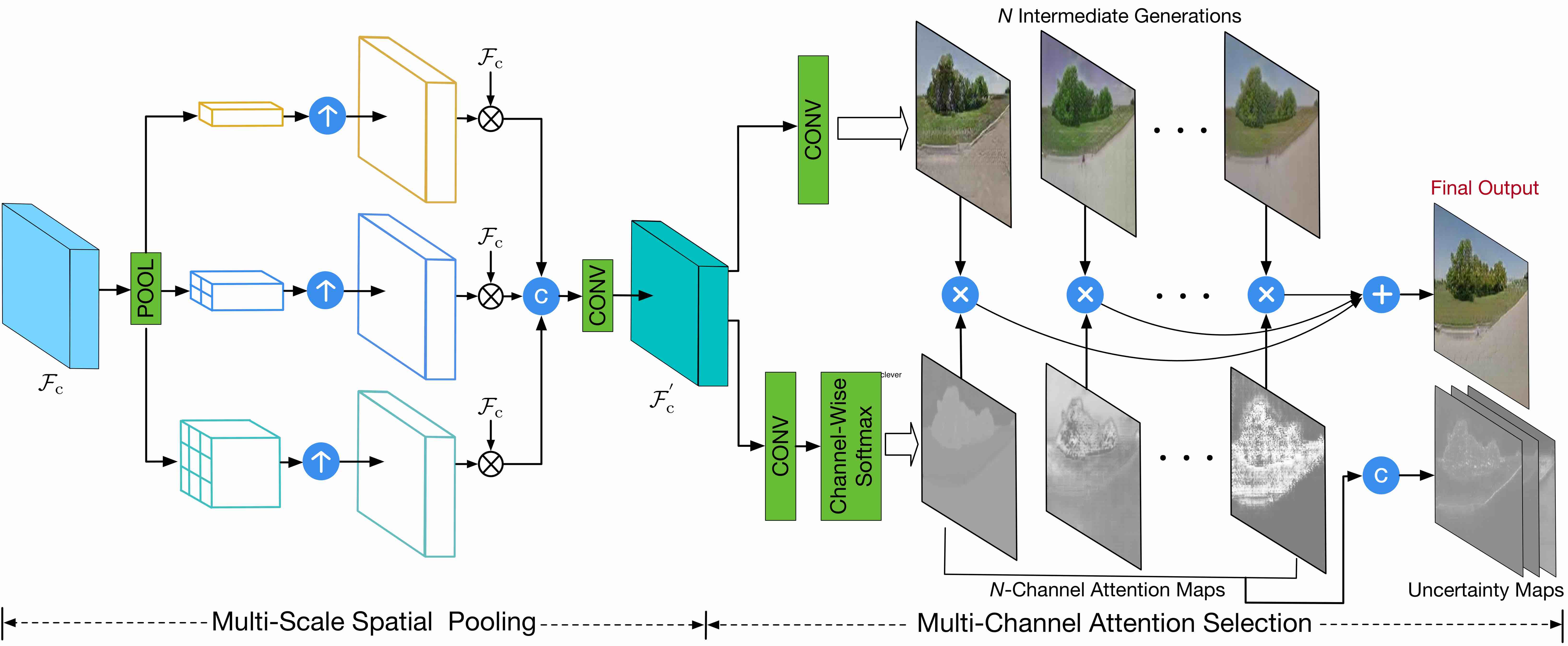}
	\caption{Illustration of the proposed multi-channel attention selection module. The multi-scale spatial pooling pools features in different receptive fields in order to have better generation of scene details; the multi-channel attention selection aims at automatically select from a set of intermediate diverse generations in a larger generation space to improve the generation quality. The symbols $\oplus$, $\otimes$, $\textcircled{c}$ and \protect\ci{$\uparrow$} denote element-wise addition, element-wise multiplication, concatenation, and up-sampling operation, respectively.}
	\label{fig:ppm}
	\vspace{-0.6cm}
\end{figure*}

\section{Multi-Channel Attention Selection GAN}
\vspace{-0.1cm}
In this section we present the details of the proposed multi-channel attention selection GAN. An illustration of the overall network structure is depicted in Fig.~\ref{fig:framework}. 
In the first stage, we present a cascade semantic-guided generation sub-network, which utilizes the images from one view and conditional semantic maps from another view as inputs, and reconstruct images in another view. These images are further input into a semantic generator to recover the input semantic map forming a generation cycle. In the second stage, the coarse synthesis and the deep features from the first stage are combined, and then are passed to the proposed multi-channel attention selection module, which aims at producing more fine-grained synthesis from a larger generation space and also at generating uncertainty maps to guide multiple optimization losses.

\subsection{Cascade Semantic-guided Generation}
\vspace{-0.2cm}
\noindent \textbf{Semantic-guided Generation.}
Cross-view synthesis is a challenging task, especially when the two views have little overlapping as in our study case, which apparently leads to ambiguity issues in the generation process. To alleviate this problem, we use semantic maps as conditional guidance. Since it is always costly to obtain annotated semantic maps, following~\cite{regmi2018cross} we generate the maps using segmentation deep models pretrained from large-scale scene parsing datasets such as Cityscapes~\cite{cordts2016cityscapes}. However, \cite{regmi2018cross} uses semantic maps only in the reconstruction loss to guide the generation of semantics, which actually provides a weak guidance. Different from theirs, we apply the semantic maps not only in the output loss but also as part of the network's input. Specifically, as shown in Fig.~\ref{fig:framework}, we concatenate the input image $I_a$ from the source view and the semantic map $S_g$ from a target view, and input them into the image generator $G_i$ and synthesize the target view image $I_g^{'}$ as $I_g^{'} {=} G_i(I_a, S_g)$.
In this way, the ground-truth semantic maps provide stronger supervision to guide the cross-view translation in the deep network.

\par\noindent \textbf{Semantic-guided Cycle.}
Regmi and Borji~\cite{regmi2018cross} observed that the simultaneous generation of both the images and the semantic maps improves the generation performance. Along the same line, we propose a cycled semantic generation network to benefit more the semantic information in learning. The conditional semantic map $S_g$ together with the input image $I_a$ are input into the image generator $G_i$, and produce the synthesized image $I_g^{'}$. Then $I_g^{'}$ is further fed into the semantic generator $G_s$ which reconstructs a new semantic map $S_g^{'}$. We can formalize the process as $S_g^{'} {=} G_s(I_g^{'}){=}G_s(G_i(I_a, S_g))$. 
Then the optimization objective is to make $S_g^{'}$ as close as possible to $S_g$, which naturally forms a semantic generation cycle, \ie~$[I_a, S_g] \stackrel{G_i} \rightarrow I_g^{'} \stackrel{G_s} \rightarrow S_g^{'}  {\approx} S_g$. The two generators are explicitly connected by the ground-truth semantic maps, which in this way provide extra constraints on the generators to learn better the semantic structure consistency. 

\noindent \textbf{Cascade Generation.} Due to the complexity of the task, after the first stage, we observe that the image generator $G_i$ outputs a coarse synthesis, which yields blurred scene details and high pixel-level dis-similarity with the target-view images. This inspires us to explore a coarse-to-fine generation strategy in order to boost the synthesis performance based on the coarse predictions. Cascade models have been used in several other computer vision tasks such as object detection~\cite{chen2014joint} and semantic segmentation~\cite{dai2016instance}, and have shown great effectiveness. In this paper, we introduce the cascade strategy to deal with the complex cross-view translation problem. In both stages we have a basic cycled semantic guided generation sub-network, while in the second stage, we propose a novel multi-channel attention selection module to better utilize the coarse outputs from the first stage and produce fine-grained final outputs. We observed significant improvement by using the proposed cascade strategy, illustrated in the experimental part. 

\subsection{Multi-Channel Attention Selection}
\vspace{-0.2cm}
An overview of the proposed multi-channel attention selection module $G_a$ is shown in Fig.~\ref{fig:ppm}. The module consists of a multi-scale spatial pooling and a multi-channel attention selection component. 

\noindent \textbf{Multi-Scale Spatial Pooling.}
Since there exists a large object/scene deformation between the source view and the target view, a single-scale feature may not be able to capture all the necessary spatial information for a fine-grained generation. Thus we propose a multi-scale spatial pooling scheme, which uses a set of different kernel size and stride to perform a global average pooling on the same input features. By so doing, we obtain multi-scale features with different receptive fields to perceive a different spatial context. More specifically, given the coarse inputs and the deep semantic features produced from the stage I, we first concatenate all of them as new features denoted as $\mathcal{F}_{\mathrm{c}}$ for the stage II as:
\vspace{-0.1cm}
\begin{equation}
\begin{aligned}
\mathcal{F}_{\mathrm{c}} = \mathrm{concat}(I_a, I_g^{'}, F_i, F_s)
\end{aligned}
\vspace{-0.1cm}
\end{equation}
where $\mathrm{concat}(\cdot)$ is a function for channel-wise concatenation operation; $F_i$ and $F_s$ are features from the last convolution layers of the generators $G_i$ and $G_s$, respectively. We apply a set of $M$ spatial scales $\{s_i\}_{i=1}^M$ in pooling, resulting in pooled features with different spatial resolution.
Different from the pooling scheme used in~\cite{zhao2017pyramid} which directly combines all the features after pooling, we first select each pooled feature via an element-wise multiplication with the input feature. 
Since in our task the input features are from different sources, 
highly correlated features would preserve more useful information for the generation. Let us denote $\mathrm{pl\_up_s}(\cdot)$ as pooling at a scale $s$ followed by an up-sampling operation to rescale the pooled feature at the same resolution, and $\otimes$ as element-wise multiplication, we can formalize the whole process as follows:
\vspace{-0.1cm}
\begin{equation}
\begin{aligned}
\mathcal{F}_{\mathrm{c}} \leftarrow \mathrm{concat}\big(\mathcal{F}_{\mathrm{c}} \otimes \mathrm{pl\_up_1}(\mathcal{F}_{\mathrm{c}}),\dots, \mathcal{F}_{\mathrm{c}} \otimes \mathrm{pl\_up}_M(\mathcal{F}_{\mathrm{c}}))
\end{aligned}
\vspace{-0.1cm}
\end{equation}
Then the features $\mathcal{F}_{\mathrm{c}}$ are fed into a convolutional layer, which produces new multi-scale features $\mathcal{F}_{\mathrm{c}}^{'}$ for the use in the multi-channel selection module. 

\noindent \textbf{Multi-Channel Attention Selection.} 
In previous cross-view image synthesis works, the image is generated only in a three-channel RGB space. We argue that this is not enough for the complex translation problem we are dealing with, and thus we explore using a larger generation space to have a richer synthesis via constructing multiple intermediate generations.
Accordingly, we design a multi-channel attention mechanism to automatically perform spatial and temporal selection from the generations to synthesize a fine-grained final output.

\par Given the multi-scale feature volume $\mathcal{F}_{\mathrm{c}}^{'}{\in} \mathbb{R}^{h\times w \times c}$, where $h$ and $w$ are width and height of the features, and $c$ is the number of channels, we consider two directions. One is for the generation of multiple intermediate image synthesis, and the other is for the generation of multi-channel attention maps. To produce $N$ different intermediate generations $I_G{=}\{I_G^i\}_{i=1}^N$, a convolution operation is performed with $N$ convolutional filters $\{W_G^i,b_G^i\}_{i=1}^N$ followed by a $\tanh(\cdot)$ non-linear activation operation. For the generation of corresponding $N$ attention maps, the other group of filters $\{W_A^i,b_A^i\}_{i=1}^N$ is applied. Then the intermediate generations and the attention maps are calculated as follows:
\vspace{-0.1cm}
\begin{equation} 
\begin{aligned}
I_G^i = \tanh(\mathcal{F}_{\mathrm{c}}^{'}W_G^i + b_G^i), \quad \quad \text{for} \ \ i = 1, \dots, N \\
I_A^i = \mathrm{Softmax}(\mathcal{F}_{\mathrm{c}}^{'}W_A^i + b_A^i), \quad \quad \text{for} \ \ i = 1, \dots, N 
\end{aligned}
\label{eqn:attention}
\vspace{-0.1cm}
\end{equation}
where $\mathrm{Softmax}(\cdot)$ is a channel-wise softmax function used for the normalization. Finally, the learned attention maps are utilized to perform channel-wise selection from each intermediate generation as follows:
\vspace{-0.1cm}
\begin{equation}
\begin{aligned}
I_g^{''} = (I_A^1 \otimes I_G^1) \oplus \cdots \oplus (I_A^N \otimes I_G^N) 
\end{aligned}
\vspace{-0.1cm}
\end{equation}
where $I_g^{''}$ represents the final synthesized generation selected from the multiple diverse results, and the symbol $\oplus$ denotes the element-wise addition.
We also generate a final semantic map in the second stage as in the first stage, \ie $
S_g^{''}  {=} G_s(I_g^{''})$. Due to the same purpose of the two semantic generators, we use a single $G_s$ twice by sharing the parameters in both stages to reduce the network capacity.

\noindent \textbf{Uncertainty-guided Pixel Loss.}
As we discussed in the introduction, the semantic maps obtained from the pretrained model are not accurate for all the pixels, which leads to a wrong guidance during training. To tackle this issue, we propose the generated attention maps to learn uncertainty maps to control the optimization loss. The uncertainty learning has been investigated in~\cite{kendall2017multi} for multi-task learning, and here we introduce it for solving the noisy semantic label problem. 
Assume that we have $K$ different loss maps which need a guidance. The multiple generated attention maps are first concatenated and passed to a convolution layer with $K$ filters $\{W_u^i\}_{i=1}^K$ to produce a set of $K$ uncertainty maps. The reason of using the attention maps to generate uncertainty maps is that the attention maps directly affect the final generation leading to a close connection with the loss. Let $\mathcal{L}_p^i$ denote a pixel-level loss map and $U_i$ denote the $i$-th uncertainty map, we have:
\vspace{-0.1cm}
\begin{equation}
\begin{aligned}
&U_i = \sigma\big(W_u^i(\mathrm{concat}(I_A^1,\dots,I_A^N) + b_u^i\big) \\
&\mathcal{L}_{p}^i \leftarrow \frac{\mathcal{L}_{p}^i }{U_i}+ \log U_i, \quad \text{for} \ \  i=1,\dots,K
\end{aligned}
\vspace{-0.1cm}
\end{equation}
where $\sigma(\cdot)$ is a Sigmoid function for pixel-level normalization. The uncertainty map is automatically learned and acts as a weighting scheme to control the optimization loss.

\noindent \textbf{Parameter-Sharing Discriminator.}
We extend the vanilla discriminator in~\cite{isola2017image} to a parameter-sharing structure. 
In the first stage, this structure takes the real image $I_a$ and the generated image $I_g^{'}$ or the ground-truth image $I_g$ as input. The discriminator $D$ learns to tell whether a pair of images from different domains is associated with each other or not. In the second stage, it accepts the real image $I_a$ and the generated image $I_g^{''}$ or the real image $I_g$ as input. This pairwise input encourages $D$ to discriminate the diversity of image structure and capture the local-aware information. 

\subsection{Overall Optimization Objective}
\vspace{-0.2cm}
\noindent \textbf{Adversarial Loss.}
In the first stage, the adversarial loss of $D$ for distinguishing synthesized image pairs $[I_a, I_g^{'}]$ from real image pairs $[I_a, I_g]$ is formulated as follows,
\vspace{-0.1cm}
\begin{equation}
\begin{aligned}
\mathcal{L}_{cGAN}(I_a, I_g^{'}) = 
& \mathbb{E}_{I_a, I_g} \left[ \log D(I_a, I_g) \right] +  \\
& \mathbb{E}_{I_a, I_g^{'}} \left[\log (1 - D(I_a, I_g^{'})) \right].
\end{aligned}
\label{eqn:adv_1}
\vspace{-0.1cm}
\end{equation}
In the second stage, the adversarial loss of $D$ for distinguishing synthesized image pairs $[I_a, I_g^{''}]$ from real image pairs $[I_a, I_g]$ is formulated as follows:
\vspace{-0.1cm}
\begin{equation}
\begin{aligned}
\mathcal{L}_{cGAN}(I_a, I_g^{''}) {=} 
& \mathbb{E}_{I_a, I_g} \left[ \log D(I_a, I_g) \right] +  \\
& \mathbb{E}_{I_a, I_g^{''}} \left[\log (1 - D(I_a, I_g^{''})) \right].
\end{aligned}
\label{eqn:adv_2}
\vspace{-0.1cm}
\end{equation}
Both losses aim to preserve the local structure information and produce visually pleasing synthesized images.
Thus, the adversarial loss of the proposed SelectionGAN is the sum of Eq.~\eqref{eqn:adv_1} and  \eqref{eqn:adv_2},
\vspace{-0.1cm}
\begin{equation}
\begin{aligned}
\mathcal{L}_{cGAN} =  \mathcal{L}_{cGAN}(I_a, I_g^{'}) +  \lambda \mathcal{L}_{cGAN}(I_a, I_g^{''}).
\end{aligned}
\label{eq:adv}
\vspace{-0.1cm}
\end{equation}

\begin{figure*}[!t] \small
	\centering
	\includegraphics[width=0.9\linewidth]{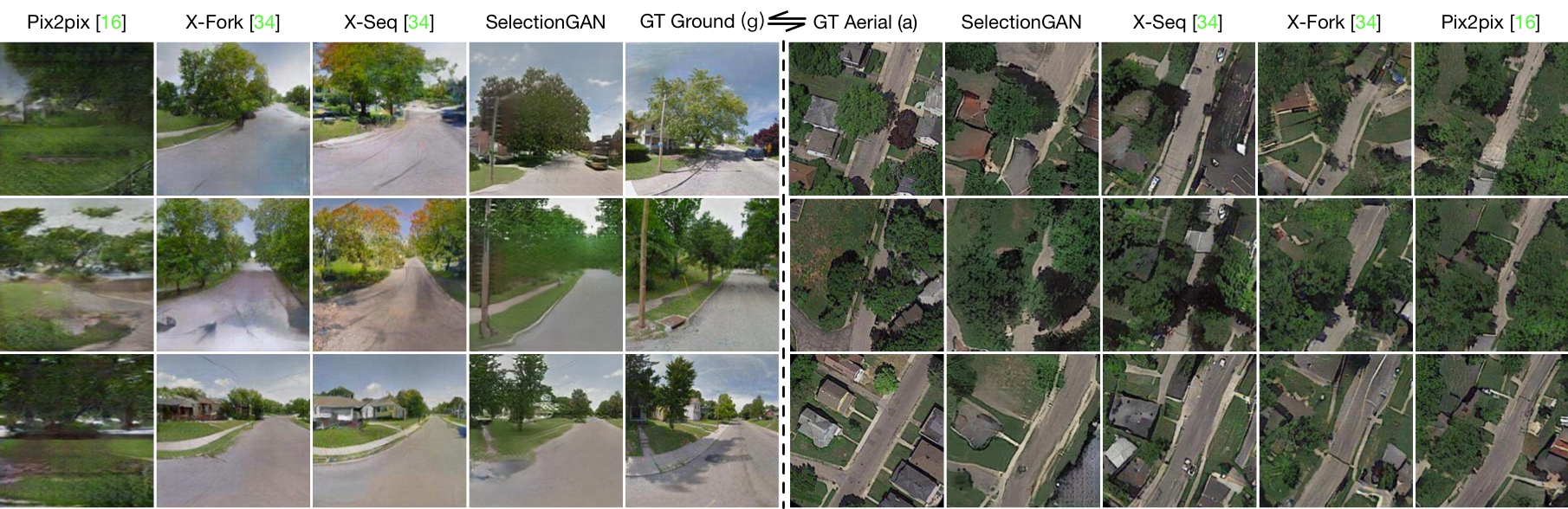}
	\caption{Results generated by different methods in $256{\times}256$ resolution in a2g and g2a directions on Dayton dataset.
	}
	\label{fig:day256}
	\vspace{-0.4cm}
\end{figure*}

\begin{table*}[!t] \small
	\centering
	\caption{SSIM, PSNR, Sharpness Difference (SD) and KL score (KL) of different methods. For these metrics except KL score, higher is better. (*) These results are reported in \cite{regmi2018cross}.}
	\resizebox{0.9\linewidth}{!}{% 
		\begin{tabular}{cccccccccccccc} \toprule
			Direction & \multirow{2}{*}{Method} & \multicolumn{4}{c}{Dayton (64$\times$64)} & \multicolumn{4}{c}{Dayton (256$\times$256)} & \multicolumn{4}{c}{CVUSA}  \\ \cmidrule(lr){3-6}	\cmidrule(lr){7-10} \cmidrule(lr){11-14} 
			$\leftrightarrows$ & &  SSIM & PSNR & SD & KL & SSIM & PSNR & SD & KL & SSIM & PSNR & SD & KL \\ \hline
			\multirow{5}{*}{a2g} & Zhai \etal \cite{zhai2017predicting} &-      & -      & -    & -      & -     & -     & -      & -      &0.4147*&17.4886*&16.6184*  & 27.43 $\pm$ 1.63* \\
			& Pix2pix \cite{isola2017image}           &0.4808*&19.4919*&16.4489*& 6.29 $\pm$ 0.80* & 0.4180*&17.6291*&19.2821*& 38.26 $\pm$ 1.88*  & 0.3923*&17.6578*&18.5239* & 59.81 $\pm$ 2.12*\\ 
			& X-Fork \cite{regmi2018cross}                &0.4921*&19.6273*&16.4928*& 3.42 $\pm$ 0.72* &0.4963*&19.8928*&19.4533*  &6.00 $\pm$ 1.28*  &0.4356*&19.0509*&18.6706* & 11.71 $\pm$ 1.55*\\
			& X-Seq \cite{regmi2018cross}               &0.5171*&20.1049*&16.6836* &6.22 $\pm$ 0.87* &0.5031*&20.2803*&19.5258* &5.93 $\pm$ 1.32*&0.4231*&18.8067*&18.4378* &15.52 $\pm$ 1.73*\\
			& SelectionGAN    (Ours)	                                        & \textbf{0.6865}  & \textbf{24.6143} &  \textbf{18.2374} & \textbf{1.70  $\pm$ 0.45}& \textbf{0.5938}  & \textbf{23.8874}  & \textbf{20.0174} & \textbf{2.74 $\pm$ 0.86}&  \textbf{0.5323} &  \textbf{23.1466}& \textbf{19.6100} &\textbf{2.96  $\pm$ 0.97}  \\ \hline
			\multirow{5}{*}{g2a} & Pix2pix \cite{isola2017image}         &0.3675*&20.5135*&14.7813* & 6.39 $\pm$ 0.90*&0.2693*&20.2177*&16.9477* &7.88 $\pm$ 1.24* &-      &-       &-  & -  \\
			& X-Fork \cite{regmi2018cross}               &0.3682*&20.6933*&14.7984* & 4.45 $\pm$ 0.84* &0.2763*&20.5978*&16.9962* &6.92 $\pm$ 1.15* &-      &-       &-   &- \\
			& X-Seq \cite{regmi2018cross}                 &0.3663*&20.4239*&14.7657* &7.20 $\pm$ 0.92* &0.2725*&20.2925*&16.9285* & 7.07 $\pm$ 1.19*&- &-      &-       &-   \\
			& SelectionGAN      (Ours)                                          & \textbf{0.5118}  &  \textbf{23.2657} & \textbf{16.2894} & \textbf{2.25 $\pm$  0.56}  & \textbf{0.3284} & \textbf{21.8066} &  \textbf{17.3817}  & \textbf{3.55  $\pm$ 0.87}&- &-      &-       &-   	\\                  	
			\bottomrule		
	\end{tabular}}
	\label{tab:ssim}
	\vspace{-0.4cm}
\end{table*}

\noindent \textbf{Overall Loss.}
The total optimization loss is a weighted sum of the above losses.
Generators $G_i$, $G_s$, attention selection network $G_a$ and discriminator $D$ are trained in an end-to-end fashion optimizing the following min-max function,
\vspace{-0.1cm}
\begin{equation}
\begin{aligned}
\min_{\{G_i, G_s, G_a\}} \max_{\{D\}} \mathcal{L} = & \sum_{i=1}^4 \lambda_i \mathcal{L}_{p}^i + \mathcal{L}_{cGAN} + \lambda_{tv}\mathcal{L}_{tv}.
\end{aligned}
\label{eqn:all}
\vspace{-0.1cm}
\end{equation} 
% where $\lambda_{pixel}$ and $\lambda_{tv}$ are the trade-off parameters.
where $\mathcal{L}_p^i$ uses the L1 reconstruction to separately calculate the pixel loss between the generated images $I_g^{'}$, $S_g^{'}$, $I_g^{''}$ and $S_g^{''}$ and the corresponding real images. 
$\mathcal{L}_{tv}$ is the total variation regularization~\cite{johnson2016perceptual} on the final synthesized image $I_g^{''}$.
$\lambda_i$ and $\lambda_{tv}$ are the trade-off parameters to control the relative importance of different objectives. 
The training is performed by solving the min-max optimization problem.

\subsection{Implementation Details}
\vspace{-0.2cm}
\noindent \textbf{Network Architecture.}
For a fair comparison, we employ U-Net \cite{isola2017image} as our generator architectures $G_i$ and $G_s$.
U-Net is a network with skip connections between a down-sampling encoder and an up-sampling decoder. 
Such architecture comprehensively retains contextual and textural information, which is crucial for removing artifacts and padding textures. Since our focus is on the cross-view image generation task, $G_i$ is more important than $G_s$.
Thus we use a deeper network for $G_i$ and a shallow network for $G_s$.
Specifically, the filters in first convolutional layer of $G_i$ and $G_s$ are 64 and 4, respectively.
For the network $G_a$, the kernel size of convolutions for generating the intermediate images and attention maps are $3{\times}3$ and $1{\times }1$, respectively.
We adopt PatchGAN~\cite{isola2017image} for the discriminator $D$.

\noindent \textbf{Training Details.}
Following \cite{regmi2018cross}, we use RefineNet \cite{lin2017refinenet} and \cite{zhou2017scene} to generate segmentation maps on Dayton and Ego2Top datasets as training data, respectively.
We follow the optimization method in~\cite{goodfellow2014generative} to optimize the proposed SelectionGAN, \ie one gradient descent step on discriminator and generators alternately.
We first train $G_i$, $G_s$, $G_a$ with $D$ fixed, and then train $D$ with $G_i$, $G_s$, $G_a$  fixed.
The proposed SelectionGAN is trained and optimized in an end-to-end fashion.
We employ  Adam \cite{kingma2014adam} with momentum terms  $\beta_1{=}0.5$ and $\beta_2{=}0.999$ as our solver.
The initial learning rate for Adam is 0.0002.
The network initialization strategy is Xavier \cite{glorot2010understanding}, weights are initialized from a Gaussian distribution with standard deviation 0.2 and mean~0.

\begin{table*}[!h] \small
	\centering
	\caption{Accuracies of different methods. For this metric, higher is better. (*) These results are reported in~\cite{regmi2018cross}.}
	\resizebox{0.9\linewidth}{!}{
		\begin{tabular}{cccccccccccccccccc} \toprule
			Dir. & \multirow{2}{*}{Method} & \multicolumn{4}{c}{Dayton (64$\times$64)} & \multicolumn{4}{c}{Dayton (256$\times$256)} & \multicolumn{4}{c}{CVUSA} \\ \cmidrule(lr){3-6}	\cmidrule(lr){7-10} \cmidrule(lr){11-14} 
			$\leftrightarrows$ & &  \multicolumn{2}{c}{Top-1} & \multicolumn{2}{c}{Top-5} &\multicolumn{2}{c}{Top-1} & \multicolumn{2}{c}{Top-5} & \multicolumn{2}{c}{Top-1} & \multicolumn{2}{c}{Top-5}  \\ 
			&& \multicolumn{2}{c}{Accuracy (\%)} & \multicolumn{2}{c}{Accuracy (\%)}  & \multicolumn{2}{c}{Accuracy (\%)}  & \multicolumn{2}{c}{Accuracy (\%)}  & \multicolumn{2}{c}{Accuracy (\%)}  & \multicolumn{2}{c}{Accuracy (\%)} \\ \hline
			\multirow{5}{*}{a2g} & Zhai \etal \cite{zhai2017predicting} &-      & -      & -      & -     & -      & -  &-      & -    &13.97*&14.03*&42.09*&52.29* \\
			& Pix2pix \cite{isola2017image}              &7.90* &15.33*&27.61*&39.07*&6.80* &9.15* &23.55*&27.00*&7.33* &9.25* &25.81*&32.67*  \\ 
			& X-Fork \cite{regmi2018cross}               &16.63*&34.73*&46.35*&70.01*&30.00*&48.68*&61.57*&78.84*&20.58*&31.24*&50.51*&63.66* \\
			& X-Seq \cite{regmi2018cross}               &4.83* &5.56* &19.55*&24.96*&30.16*&49.85*&62.59*&80.70*&15.98*&24.14*&42.91*&54.41* \\
			& SelectionGAN     (Ours)	                                        & \textbf{45.37 }& \textbf{79.00}  &\textbf{83.48} &\textbf{97.74}  & \textbf{42.11}& \textbf{68.12} & \textbf{77.74}&\textbf{92.89}  & \textbf{41.52} & \textbf{65.51} &\textbf{74.32} &\textbf{89.66} \\ \hline
			\multirow{4}{*}{g2a} & Pix2pix \cite{isola2017image} &1.65*&2.24*&7.49*&12.68*&10.23*&16.02*&30.90*&40.49*& - & -& -& -\\
			& X-Fork \cite{regmi2018cross}               &4.00*&16.41*&15.42*&35.82*&10.54*&15.29*&30.76*&37.32*& - & -  &- \\
			& X-Seq \cite{regmi2018cross}                &1.55*&2.99*&6.27*&8.96*&12.30*&19.62*&35.95*&45.94* & - & - & - & - \\
			& SelectionGAN (Ours)                                                    & \textbf{14.12}&\textbf{51.81}&\textbf{39.45}&\textbf{74.70}&\textbf{20.66} & \textbf{33.70}& \textbf{51.01}  &\textbf{63.03}   &-  & -   &-  &-    \\                    	
			\bottomrule		
	\end{tabular}}
	\label{tab:accuracies}
	\vspace{-0.4cm}
\end{table*}

\begin{table*}[!ht] \small
	\centering
	\caption{Inception Score of different methods. For this metric, higher is better. (*) These results are reported in \cite{regmi2018cross}.}
	\resizebox{0.9\linewidth}{!}{
		\begin{tabular}{ccccccccccc} \toprule
			Dir. & \multirow{2}{*}{Method} & \multicolumn{3}{c}{Dayton (64$\times$64)} & \multicolumn{3}{c}{Dayton (256$\times$256)} & \multicolumn{3}{c}{CVUSA} \\ \cmidrule(lr){3-5}	\cmidrule(lr){6-8} \cmidrule(lr){9-11}
			$\leftrightarrows$ &                                                                               &  \tht{c}{all\\classes} & \tht{c}{Top-1\\class} & \tht{c}{Top-5\\classes} & \tht{c}{all\\classes} & \tht{c}{Top-1\\class} & \tht{c}{Top-5\\classes}  & \tht{c}{all\\classes} & \tht{c}{Top-1\\class} & \tht{c}{Top-5\\classes}  \\ \hline
			\multirow{7}{*}{a2g} & Zhai \etal \cite{zhai2017predicting} &-           & -         & -          & -         & -         & -          &1.8434*&1.5171* &1.8666*   \\
			& Pix2pix \cite{isola2017image}                                          &1.8029*&1.5014*&1.9300*&2.8515*&1.9342*&2.9083*&3.2771*&2.2219*&3.4312*   \\ 
			& X-Fork \cite{regmi2018cross}                                      &1.9600*&1.5908*&2.0348*&\textbf{3.0720}*&2.2402*&3.0932*&3.4432*&2.5447*&3.5567*\\
			& X-Seq \cite{regmi2018cross}                                         & 1.8503*&1.4850*&1.9623*&2.7384*&2.1304*&2.7674*&\textbf{3.8151}*&2.6738*&\textbf{4.0077}*  \\
			& SelectionGAN (Ours)	                                                                 & \textbf{2.1606}  & \textbf{1.7213} & \textbf{2.1323} &3.0613   & \textbf{2.2707} & \textbf{3.1336} &3.8074& \textbf{2.7181} &3.9197\\ \cline{2-11}
			& Real Data                                                                                           &2.3534&1.8135&2.3250 &3.8319&2.5753&3.9222&4.8741&3.2959&4.9943\\ \hline
			\multirow{5}{*}{g2a} & Pix2pix \cite{isola2017image}          &1.7970*&1.3029*&1.6101*&3.5676*&2.0325*&2.8141* &-      &-       &-       \\
			& X-Fork \cite{regmi2018cross}                                        &1.8557*&1.3162*&1.6521*&3.1342*&1.8656*&2.5599* &-      &-       &-        \\
			& X-Seq \cite{regmi2018cross}                                          &1.7854*&1.3189*&1.6219*&\textbf{3.5849}*&2.0489*&2.8414*&-      &-       &-       \\
			& SelectionGAN   (Ours)                                                                      & \textbf{2.1571} & \textbf{1.4441} & \textbf{2.0828}  & 3.2446 & \textbf{2.1331} & \textbf{3.4091} &-      &-       &-      	\\ \cline{2-11}
			& Real Data                                                                                           &2.3015&1.5056&2.2095&3.7196&2.3626&3.8998 &-      &-       &-      \\   	                  	
			\bottomrule		
	\end{tabular}}
	\label{tab:is}
	\vspace{-0.35cm}
\end{table*}

\begin{table}[!ht] \small
	\centering
	\caption{Ablations study of the proposed SelectionGAN.} 
	\resizebox{0.9\linewidth}{!}{
		\begin{tabular}{c|c|c|c|c} \toprule
			Baseline   & Setup         & SSIM   & PSNR      & SD                                 \\	\hline	
			A	          & $I_a \stackrel{G_i} \rightarrow I_g^{'}$ & 0.4555  & 19.6574 & 18.8870                       \\ \hline
			B & $ S_g \stackrel{G_i} \rightarrow I_g^{'}$  & 0.5223 & 22.4961 & 19.2648 \\ \hline
			C	          & $[I_a, S_g] \stackrel{G_i} \rightarrow I_g^{'}$ & 0.5374  & 22.8345 & 19.2075              \\ \hline
			D             & $[I_a, S_g] \stackrel{G_i} \rightarrow I_g^{'} \stackrel{G_s} \rightarrow S_g^{'}$ & 0.5438 & 22.9773 & 19.4568 \\ \hline
			E             & \tht{c}{D + Uncertainty-Guided Pixel Loss} &  0.5522 & 23.0317 & 19.5127 \\ \hline
			F             & \tht{c}{E + Multi-Channel Attention Selection} & 0.5989 & 23.7562 & 20.0000 \\ \hline
			G            & \tht{c}{F + Total Variation Regularization} & 0.6047  & 23.7956 & 20.0830  \\ \hline
			H            & \tht{c}{G + Multi-Scale Spatial Pooling} & \textbf{0.6167} & \textbf{23.9310} & \textbf{20.1214} \\ 
			\bottomrule		
	\end{tabular}}
	\label{tab:ablation}
\end{table}
\vspace{-0.1cm}
\section{Experiments}
\vspace{-0.1cm}
\label{sec:experiments}
\subsection{Experimental Setting}
\vspace{-0.2cm}
\noindent \textbf{Datasets.}
We perform the experiments on three different datasets: (i) For the Dayton dataset~\cite{vo2016localizing}, following the same setting of~\cite{regmi2018cross}, we select 76,048 images and create a train/test split of 55,000/21,048 pairs. 
The images in the original dataset have $354{\times}354$ resolution. 
We resize them to $256{\times}256$; (ii) The CVUSA dataset \cite{workman2015wide} consists of 35,532/8,884 image pairs in train/test split. 
Following~\cite{zhai2017predicting,regmi2018cross}, the aerial images are center-cropped to $224{\times}224$ and resized to $256{\times}256$. 
For the ground level images and corresponding segmentation maps, we take the first quarter of both and resize them to $256{\times}256$; (iii) The Ego2Top dataset~\cite{ardeshir2016ego2top} is more challenging and contains different indoor and outdoor conditions.
Each case contains one top-view video and several egocentric videos captured by the people visible in the top-view camera. This dataset has more than 230,000 frames. For training data, we randomly select 386,357 pairs and each pair is composed of two images of the same scene but different viewpoints.
We randomly select 25,600 pairs for evaluation.

\noindent \textbf{Parameter Settings.}
For a fair comparison, we adopt the same training setup as in~\cite{isola2017image,regmi2018cross}. All images are scaled to $256{\times}256$, and we enabled image flipping and random crops for data augmentation.
Similar to~\cite{regmi2018cross}, the low resolution ($64{\times}64$) experiments on Dayton dataset are carried out for 100 epochs with batch size of 16, whereas the high resolution ($256{\times}256$) experiments for this dataset are trained for 35 epochs with batch size of 4.
For the CVUSA dataset, we follow the same setup as in~\cite{zhai2017predicting,regmi2018cross}, and train our network for 30 epochs with batch size of 4.
For the Ego2Top dataset, all models are trained with 10 epochs using batch size 8. In our experiment, we set $\lambda_{tv}$=$1e{-}6$, $\lambda_1{=}100$, $\lambda_2{=}1$, $\lambda_3{=}200$ and $\lambda_4{=}2$ in Eq.~\eqref{eqn:all}, and $\lambda{=}4$ in Eq.~\eqref{eq:adv}.
The number of attention channels $N$ in Eq.~\eqref{eqn:attention} is set to 10.
The proposed SelectionGAN is implemented in PyTorch.
We perform our experiments on Nvidia GeForce GTX 1080 Ti GPU with 11GB memory to accelerate both training and inference.

\noindent \textbf{Evaluation Protocol.}
Similar to~\cite{regmi2018cross}, we employ Inception Score, top-k prediction accuracy and KL score for the quantitative analysis.
These metrics evaluate the generated images from a high-level feature space. We also employ pixel-level similarity metrics to evaluate our method, \emph{i.e.} Structural-Similarity (SSIM), Peak Signal-to-Noise Ratio (PSNR) and Sharpness Difference (SD).

\begin{table*}[!t]\small
	\centering
	\caption{Quantitative results on Ego2Top dataset. For all metrics except KL score, higher is better.}
	\resizebox{0.9\linewidth}{!}{ 
		\begin{tabular}{cccccccccccc} \toprule
			\multirow{2}{*}{Method}  & \multirow{2}{*}{SSIM}  & \multirow{2}{*}{PSNR} & \multirow{2}{*}{SD} & \multicolumn{3}{c}{Inception Score}                                             &\multicolumn{4}{c}{Accuracy} &  \multirow{2}{*}{KL Score}  \\   \cmidrule(lr){5-7}  \cmidrule(lr){8-11}                   
			&   &   &  & \tht{c}{all classes} & \tht{c}{Top-1 class} & \tht{c}{Top-5 classes} & \multicolumn{2}{c}{Top-1}  &  \multicolumn{2}{c}{Top-5} &  \\ \midrule	
			Pix2pix \cite{isola2017image} & 0.2213 & 15.7197 & 16.5949 & 2.5418 & 1.6797  & 2.4947  & 1.22 &1.57 &5.33 &6.86 & 120.46 $\pm$ 1.94   \\ 
			X-Fork \cite{regmi2018cross}  & 0.2740 & 16.3709 & 17.3509 &4.6447    &2.1386       &3.8417       & 5.91  & 10.22 & 20.98 & 30.29 & 22.12 $\pm$ 1.65  \\
			X-Seq \cite{regmi2018cross}   & 0.2738 & 16.3788 & 17.2624 &4.5094    &2.0276       &3.6756       & 4.78      & 8.96 & 17.04 & 24.40 & 25.19 $\pm$ 1.73  \\
			SelectionGAN (Ours)  & \textbf{0.6024} & \textbf{26.6565} & \textbf{19.7755} & \textbf{5.6200} & \textbf{2.5328} & \textbf{4.7648} & \textbf{28.31}   & \textbf{54.56} &\textbf{62.97} &\textbf{76.30} & \textbf{3.05 $\pm$ 0.91}  \\  \hline 
			Real Data                     &-      &-     &-   &6.4523     &2.8507       &5.4662       &-           &- & - & - & -  \\ 
			\bottomrule		
	\end{tabular}}
	\label{tab:ego}
	\vspace{-0.4cm}
\end{table*}

\subsection{Experimental Results}
\vspace{-0.2cm}
\noindent\textbf{Baseline Models.} 
We conduct ablation study in a2g (aerial-to-ground) direction on Dayton dataset. To reduce the training time, we randomly select 1/3 samples from the whole 55,000/21,048 samples \emph{i.e.} around 18,334 samples for training and 7,017 samples for testing. The proposed SelectionGAN considers eight baselines (A, B, C, D, E, F, G, H) as shown in Table~\ref{tab:ablation}. 
Baseline A uses a Pix2pix structure~\cite{isola2017image} and generates $I_g^{'}$ using a single image $I_a$. 
Baseline B uses the same Pix2pix model and generates $I_g^{'}$ using the corresponding semantic map $S_g$.
Baseline C also uses the Pix2pix structure, and inputs the combination of a conditional image $I_a$ and the target semantic map $S_g$ to the generator $G_i$.
Baseline D uses the proposed cycled semantic generation upon Baseline C.
Baseline E represents the pixel loss guided by the learned uncertainty maps.
Baseline F employs the proposed multi-channel attention selection module to generate multiple intermediate generations, and to make the neural network attentively select which part is more important for generating a scene image with a new viewpoint. 
Baseline G adds the total variation regularization on the final result $I_g^{''}$.
Baseline H employs the proposed multi-scale spatial pooling module to refine the features $\mathcal{F}_{\mathrm{c}}$ from stage I. All the baseline models are trained and tested on the same data using the configuration.

\begin{figure}[!t] \small
	\centering
	\includegraphics[width=0.9\linewidth]{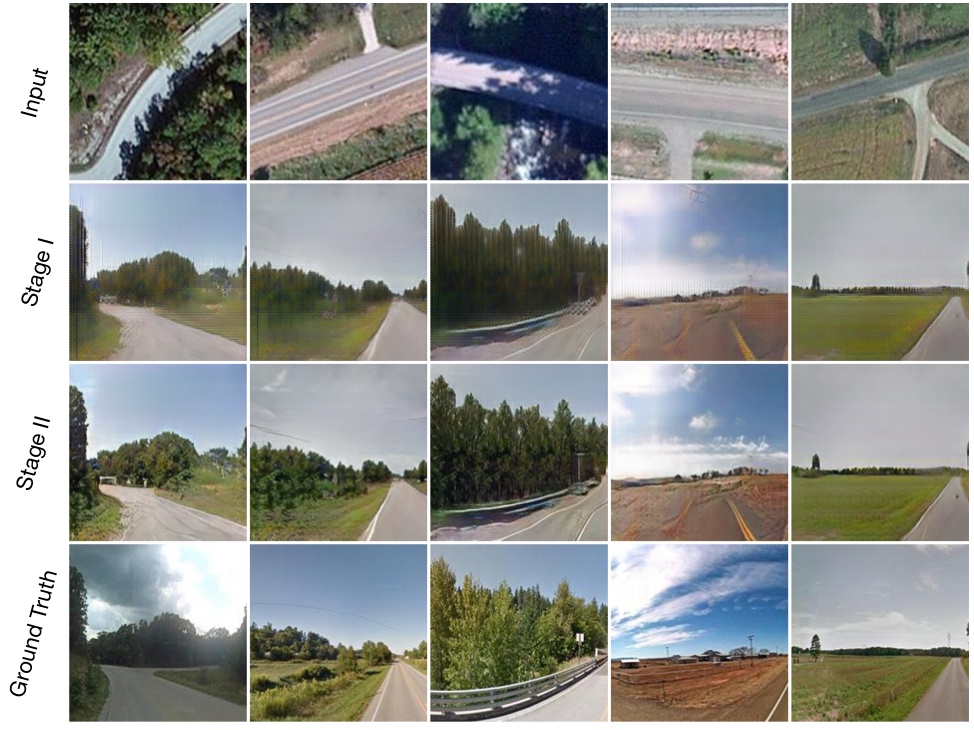}
	\caption{Qualitative results of coarse-to-fine generation on CVUSA dataset.
	}
	\label{fig:cvusa1}
\end{figure}

\noindent \textbf{Ablation Analysis.}
The results of ablation study are shown in Table~\ref{tab:ablation}. We observe that Baseline B is better than baseline A since $S_g$ contains more structural information than $I_a$. By comparison Baseline A with Baseline C, the semantic-guided generation improves SSIM, PSNR and SD by 8.19, 3.1771 and 0.3205, respectively, which confirms the importance of the conditional semantic information;
By using the proposed cycled semantic generation, Baseline D further improves over C, meaning that the proposed semantic cycle structure indeed utilizes the semantic information in a more effective way, confirming our design motivation;
Baseline E outperforms D showing the importance of using the uncertainty maps to guide the pixel loss map which contains an inaccurate reconstruction loss due to the wrong semantic labels produced from the pretrained segmentation model;
Baseline F significantly outperforms E with around 4.67 points gain on the SSIM metric, clearly demonstrating the effectiveness of the proposed multi-channel attention selection scheme;
We can also observe from Table~\ref{tab:ablation} that, by adding the proposed multi-scale spatial pool scheme and the TV regularization, the overall performance is further boosted.
Finally, we demonstrate the advantage of the proposed two-stage strategy over the one-stage method. Several examples are shown in Fig.~\ref{fig:cvusa1}. It is obvious that the coarse-to-fine generation model is able to generate sharper results and contains more details than the one-stage model.

\begin{figure}[!t] \small
	\centering
	\includegraphics[width=0.9\linewidth]{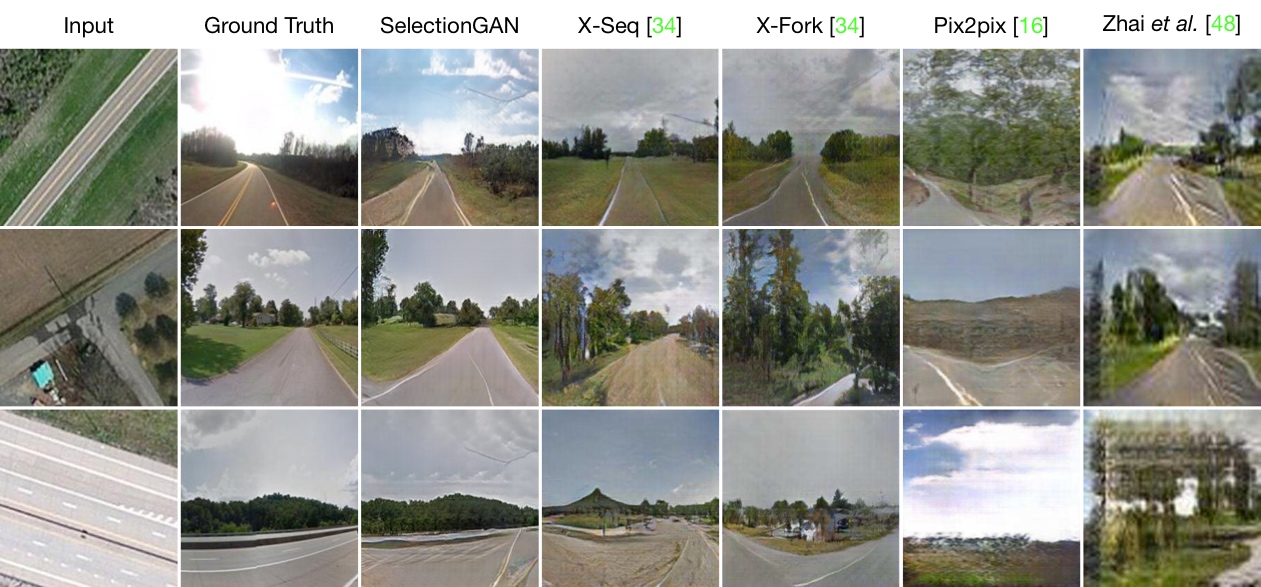}
	\caption{Results generated by different methods in $256{\times}256$ resolution in a2g direction on CVUSA dataset.
	}
	\label{fig:cvusa_comparsion}
\end{figure}

\noindent \textbf{State-of-the-art Comparisons.}
We compare our SelectionGAN with four recently proposed state-of-the-art methods, which are Pix2pix~\cite{isola2017image}, Zhai~\emph{et al.}~\cite{zhai2017predicting}, X-Fork~\cite{regmi2018cross} and X-Seq~\cite{regmi2018cross}.
The comparison results are shown in Tables~\ref{tab:ssim},~\ref{tab:accuracies},~\ref{tab:is}, and~\ref{tab:ego}.
We can observe the significant improvement of SelectionGAN in these tables. SelectionGAN consistently outperforms Pix2pix, Zhai~\emph{et al.}, X-Fork and X-Seq on all the metrics except for Inception Score. In some cases in Table~\ref{tab:is} we achieve a slightly lower performance as compared with X-Seq. However, we generate much more photo-realistic results than X-Seq as shown in Fig.~\ref{fig:day256} and~\ref{fig:cvusa_comparsion}.

\noindent \textbf{Qualitative Evaluation.}
The qualitative results in higher resolution on Dayton and CVUSA datasets are shown in Fig.~\ref{fig:day256} and~\ref{fig:cvusa_comparsion}.
It can be seen that our method generates more clear details on objects/scenes such as road, tress, clouds, car than the other comparison methods in the generated ground level images. For the generated aerial images, we can observe that grass, trees and house roofs are well rendered compared to others.
Moreover, the results generated by our method are closer to the ground truths in layout and structure, such as the results in a2g direction in Fig.~\ref{fig:day256} and~\ref{fig:cvusa_comparsion}.

\begin{figure}[!t]\small
	\centering
	\includegraphics[width=0.88\linewidth]{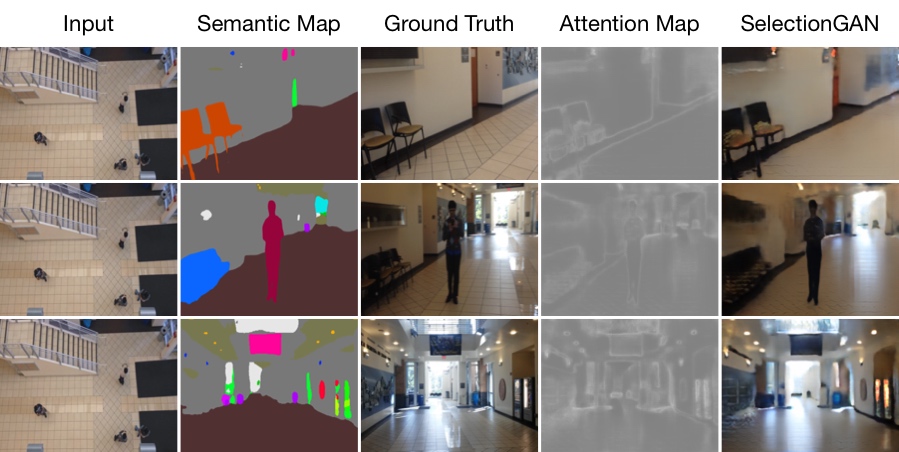}
	\caption{Arbitrary cross-view image translation on Ego2Top dataset.
	}
	\label{fig:ego}
\end{figure}

\noindent \textbf{Arbitrary Cross-View Image Translation.}
Since Dayton and CVUSA datasets only contain two views in one scene, \emph{i.e.} aerial and ground views. We further use the Ego2Top dataset to conduct the arbitrary cross-view image translation experiments. The quantitative and qualitative results are shown in Table~\ref{tab:ego} and Fig.~\ref{fig:ego}, respectively. Given an image and some novel semantic maps, SelectionGAN is able to generate the same scene but with different viewpoints.

\vspace{-0.3cm}
\section{Conclusion}
\vspace{-0.2cm}
We propose the Multi-Channel Attention Selection GAN (SelectionGAN) to address a novel image synthesizing task by conditioning on a reference image and a target semantic map. In particular, we adopt a cascade strategy to divide the generation procedure into two stages. 
Stage I aims to capture the semantic structure of the scene and Stage II focus on more appearance details via the proposed multi-channel attention selection module. 
We also propose an uncertainty map-guided pixel loss to solve the inaccurate semantic labels issue for better optimization.
Extensive experimental results on three public datasets demonstrate that our method obtains much better results than the state-of-the-art. 

\small\noindent\textbf{Acknowledgements:} This research was partially supported by National Institute of Standards and Technology Grant 60NANB17D191 (YY, JC),  Army Research Office W911NF-15-1-0354 (JC) and gift donation from Cisco \textit{Inc} (YY).

{
\bibliographystyle{ieee}
\bibliography{egbib}
}

\clearpage

This supplementary document provides additional results
supporting the claims of the main paper. 
First, we provide detailed experimental results about the influence of the number of attention channels (Sec.~\ref{supp:1}). Additionally, we compare our two-stage model with one-stage model (Sec.~\ref{supp:2}).
We also provide the visualization results of the generated uncertainty maps (Sec.~\ref{supp:3}) and the arbitrary cross-view image translation experiments on Ego2Top dataset \cite{ardeshir2016ego2top} (Sec.~\ref{supp:4}). 
Finally, we compare our SelectionGAN with the state-of-the-arts methods, \emph{i.e.} Pix2pix~\cite{isola2017image}, X-Fork~\cite{regmi2018cross} and X-Seq~\cite{regmi2018cross}.
Specifically, we compare the results of the generated segmentation maps (Sec.~\ref{supp:5}), and visualize the comparison results on Dayton~\cite{vo2016localizing}, CVUSA \cite{workman2015wide} and Ego2Top \cite{ardeshir2016ego2top} datasets (Sec.~\ref{supp:6}).

\section{Influence of the Number of Attention Channels $N$}
\label{supp:1}
We investigate the influence of the number of attention channels $N$ in Equation 3 in the main paper.
Results are shown in Table \ref{tab:attention_n}.
We observe that the performance tends to be stable after $N=10$.
Thus, taking both performance and training speed into consideration, we have set $N=10$ in all our experiments.

\begin{table}[!ht] \small
	\centering
	\caption{Influence of the number of attention channels $N$.} 
	\begin{tabular}{c|c|c|c} \toprule
		$n$   &  SSIM    & PSNR    & SD \\	\hline	
		0      & 0.5438 & 22.9773 & 19.4568  \\
		1      & 0.5522 & 23.0317 & 19.5127 \\ 
		5     &  0.5901  & 23.8068 & \textbf{20.0033} \\
		10    & \textbf{0.5986}  & 23.7336 & 19.9993 \\
		32	  & 0.5950  & \textbf{23.8265} & 19.9086 \\ 
		\bottomrule		
	\end{tabular}
	\label{tab:attention_n}
\end{table}

\section{Coarse-to-Fine Generation}
\label{supp:2}
We provide more comparison results of coarse-to-fine generation in Table \ref{tab:cos} and Figures \ref{fig:dayton_a2g}, \ref{fig:dayton_g2a} and \ref{fig:cvusa}.
We observe that our two-stage method generate much visually better results than the one-stage model, which further confirms our motivations.

\begin{table}[!t] \small
	\centering
	\caption{Results of coarse-to-fine generation. The best results are marked in \red{\textbf{blue}} color.} 
	\begin{tabular}{c|c|c|c|c|c} \toprule
		Baseline  & Stage I    & Stage II & SSIM    & PSNR    & SD           \\	\hline	F             &  $\surd$ &             & 0.5551  & 23.1919  & 19.6311  \\
		F             &               & $\surd$ & \textbf{0.5989} & \textbf{23.7562} & \textbf{20.0000}  \\ \hline
		G             &  $\surd$ &            &  0.5680   & 23.2574 & 19.7371  \\
		G             &               & $\surd$ & \textbf{0.6047}  & \textbf{23.7956} & \textbf{20.0830} \\ \hline
		H            &  $\surd$ &            &  0.5567 & 23.1545 & 19.6034    \\
		H            &               & $\surd$ &  \textbf{\red{0.6167}} & \textbf{\red{23.9310}} & \textbf{\red{20.1214}}    \\ 
		\bottomrule		
	\end{tabular}
	\label{tab:cos}
\end{table}

\section{Visualization of Uncertainty Map}
\label{supp:3}
\begin{figure*}[!ht] \small
	\centering
	\includegraphics[width=1\linewidth]{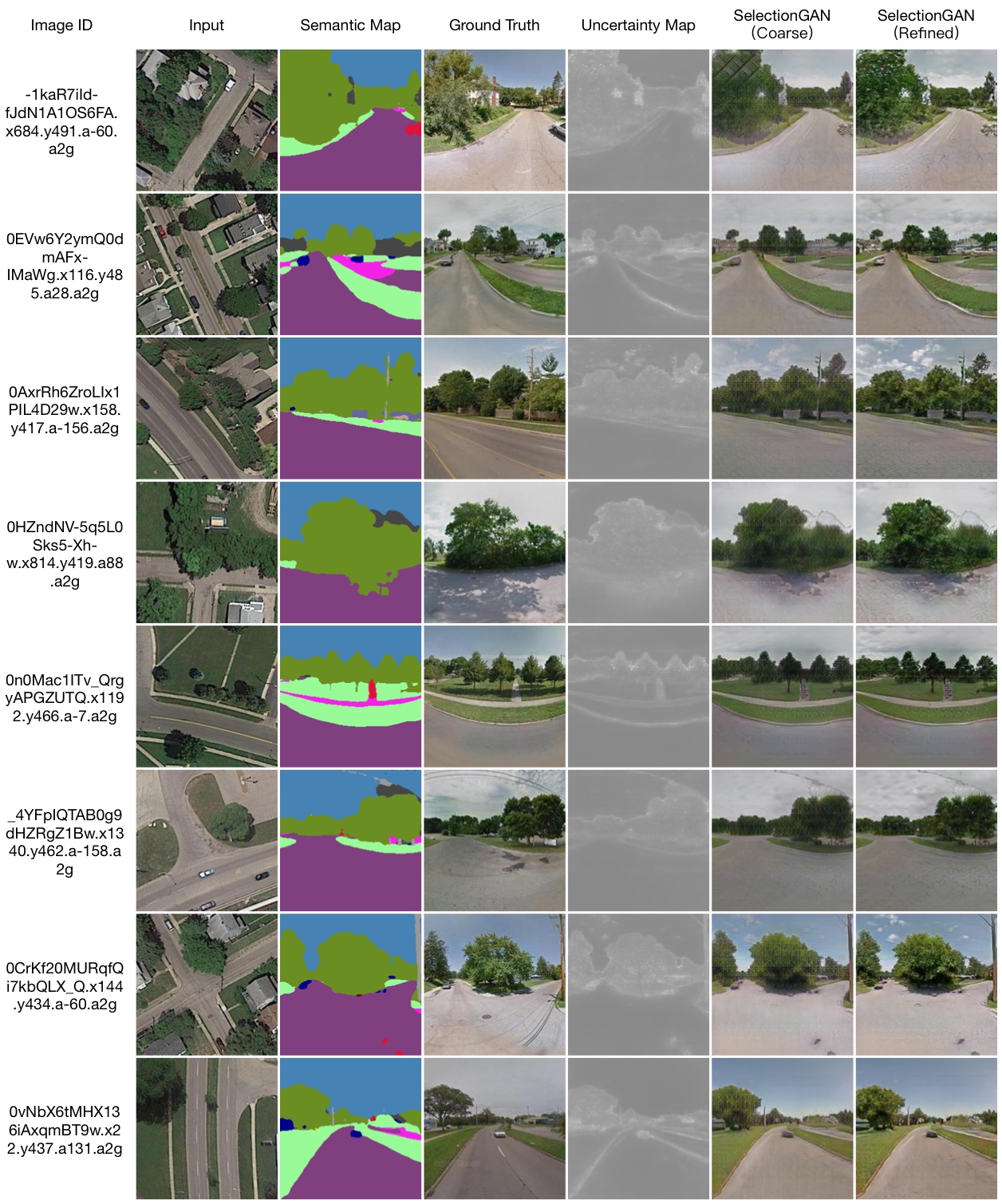}
	\caption{Results generated by our SelectionGAN in $256{\times}256$ resolution in a2g direction on Dayton dataset. These samples were randomly selected for visualization purposes.
	}
	\label{fig:dayton_a2g}
\end{figure*}

\begin{figure*}[!ht] \small
	\centering
	\includegraphics[width=1\linewidth]{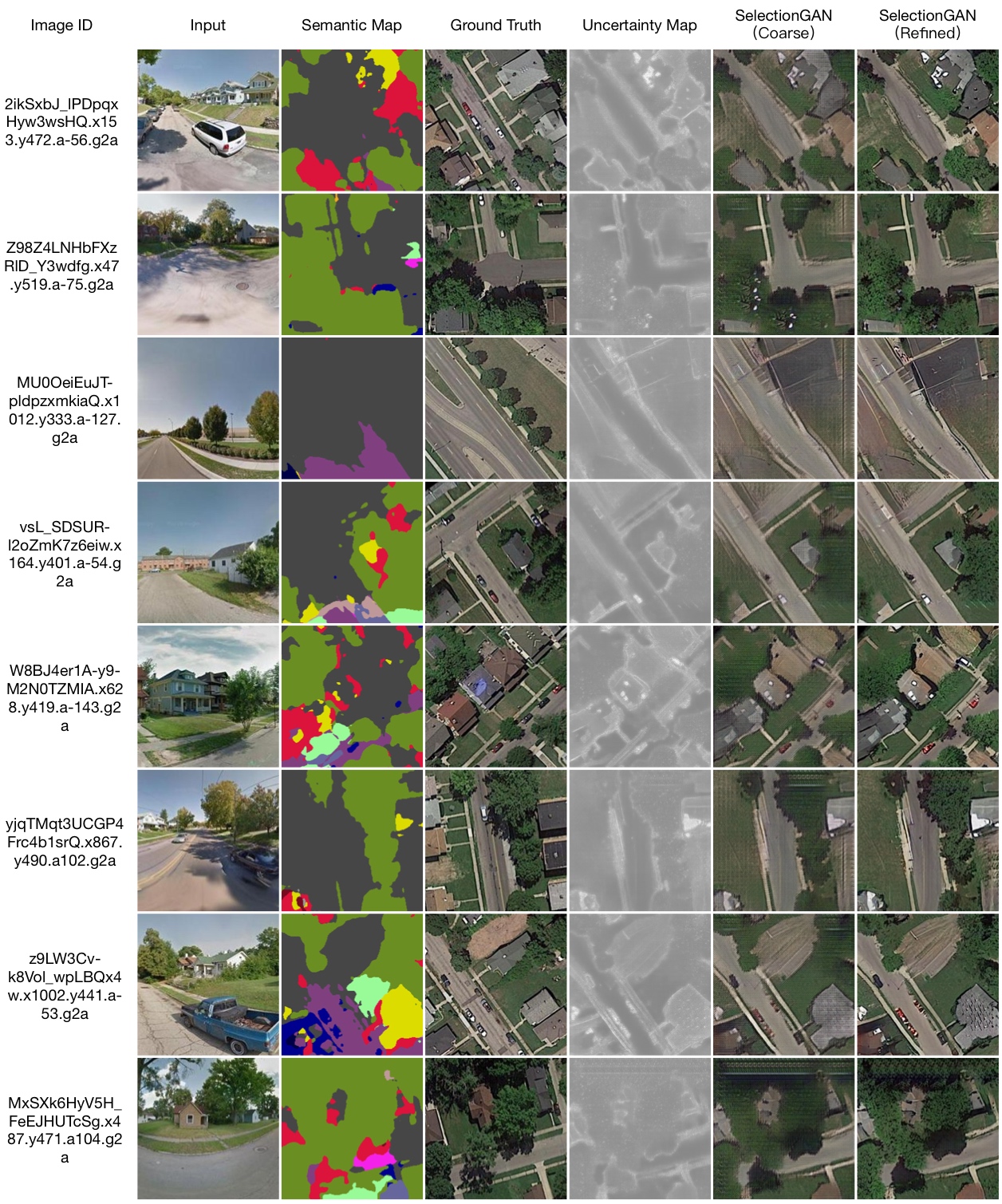}
	\caption{Results generated by our SelectionGAN in $256{\times}256$ resolution in g2a direction on Dayton dataset. These samples were randomly selected for visualization purposes.
	}
	\label{fig:dayton_g2a}
\end{figure*}

\begin{figure*}[!ht] \small
	\centering
	\includegraphics[width=1\linewidth]{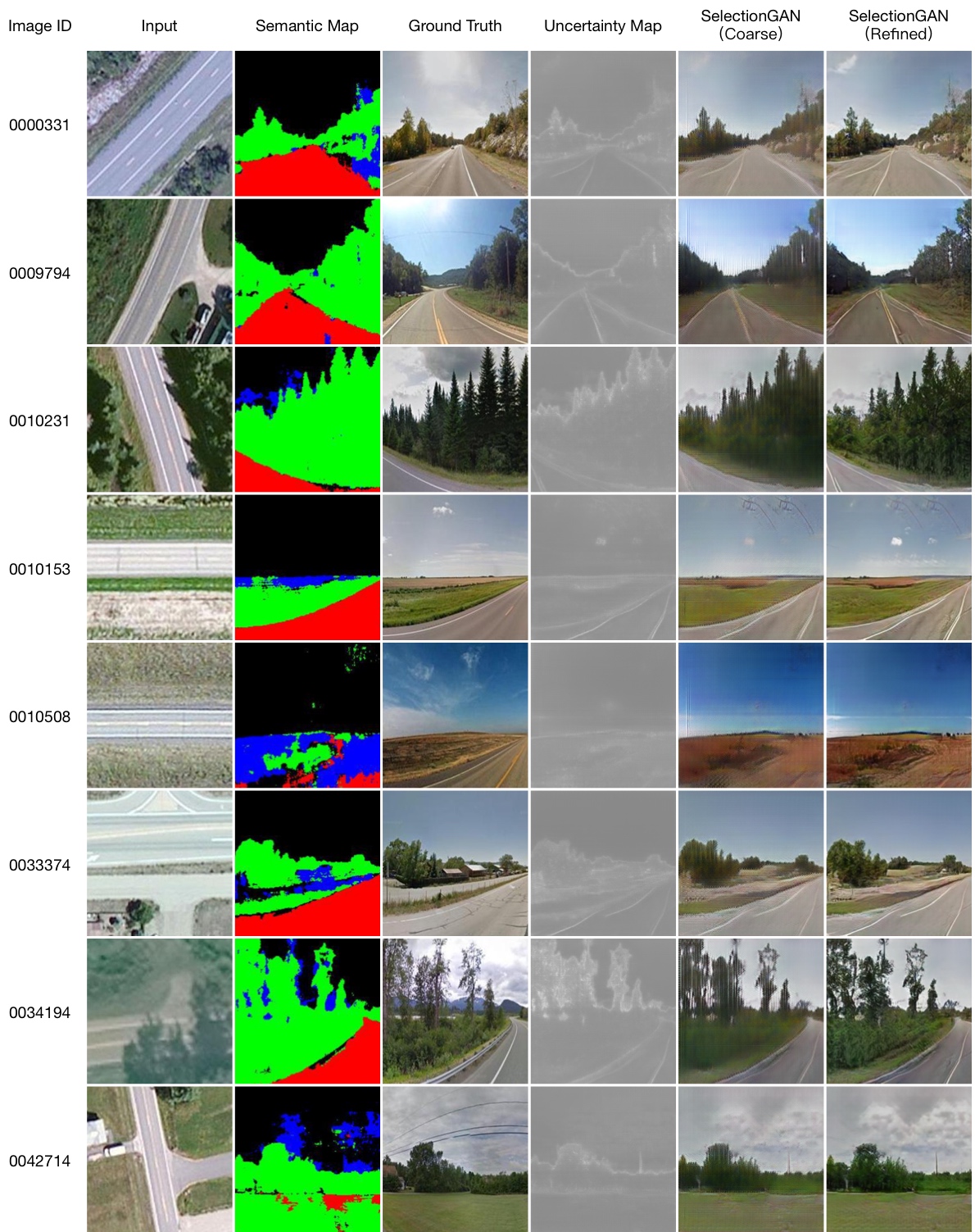}
	\caption{Results generated by our SelectionGAN in $256{\times}256$ resolution in a2g direction on CVUSA dataset. These samples were randomly selected for visualization purposes.
	}
	\label{fig:cvusa}
\end{figure*}

In Figures \ref{fig:dayton_a2g}, \ref{fig:dayton_g2a}, \ref{fig:cvusa} and  \ref{fig:ego2top_arbitrary}, we show some samples of the generated uncertainty maps.
We can see that the generated uncertainty maps learn the layout and structure of the target images.
Note that most textured regions are similar in our generation images, while the junction/edge of different regions is uncertain, and thus the model learns to highlight these parts.

\section{Arbitrary Cross-View Image Translation}
\label{supp:4}
\begin{figure*}[!ht] \small
	\centering
	\includegraphics[width=0.8\linewidth]{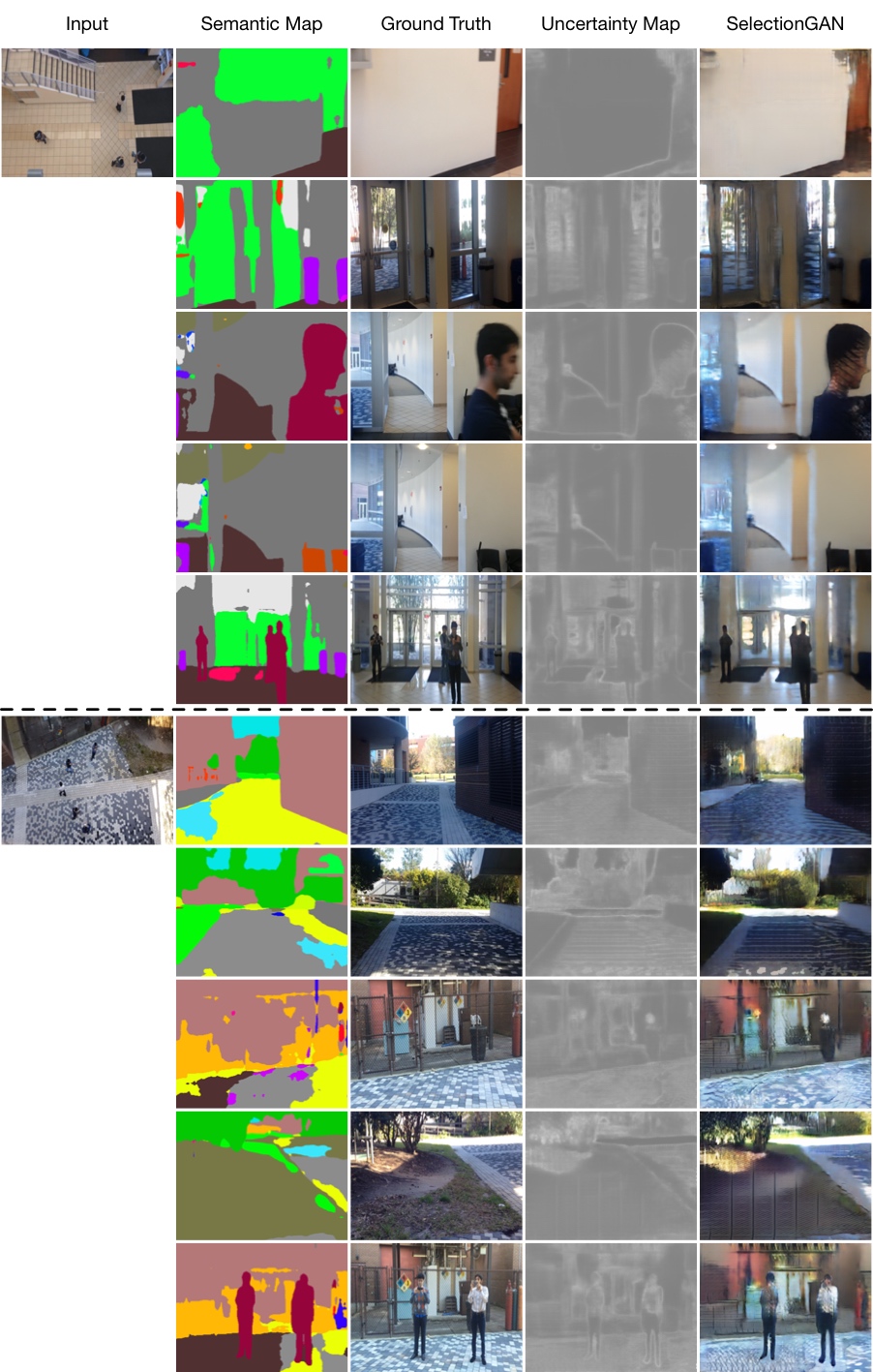}
	\caption{Arbitrary cross-view image translation on Ego2Top dataset.
	}
	\label{fig:ego2top_arbitrary}
\end{figure*}

We also conducted the arbitrary cross-view image translation experiments on Ego2Top dataset. As we can see from Figure \ref{fig:ego2top_arbitrary}, given an image and some novel semantic maps, SelectionGAN is able to generate the same scene but with different viewpoints in both outdoor and indoor environments.

\begin{table}[!ht]\small
	\centering
	\caption{Per-class accuracy and mean IOU for the generated segmentation maps on Dayton dataset. For both metric, higher is better. (*) These results are reported in \cite{regmi2018cross}.}
	\begin{tabular}{ccc} \toprule
		\multirow{2}{*}{Method}  & \multicolumn{2}{c}{a2g}  \\ \cmidrule(lr){2-3}  
		& Per-Class Acc.   & mIOU    \\ \midrule	
		X-Fork \cite{regmi2018cross}  & 0.6262*&0.4163* \\
		X-Seq \cite{regmi2018cross}   & 0.4783*&0.3187*\\
		SelectionGAN (Ours)	& \textbf{0.6415} & \textbf{0.5455}\\             	
		\bottomrule		
	\end{tabular}
	\label{tab:seg}
\end{table}

\section{Generated Segmentation Maps}
\label{supp:5}
Since the proposed SelectionGAN can generate segmentation maps, we also compare it with X-Fork~\cite{regmi2018cross} and X-Seq~\cite{regmi2018cross} on Dayton dataset.
Following \cite{regmi2018cross}, we compute per-class accuracies and mean IOU for the most common classes in this dataset: ``vegetation'', ``road'', ``building'' and ``sky'' in
ground segmentation maps.
Results are shown in Table \ref{tab:seg}.
We can see that the proposed SelectionGAN achieves better results than X-Fork~\cite{regmi2018cross} and X-Seq~\cite{regmi2018cross} on both metrics.

\section{State-of-the-art Comparisons}
\label{supp:6}
\begin{figure*}[!ht] \small
	\centering
	\includegraphics[width=1\linewidth]{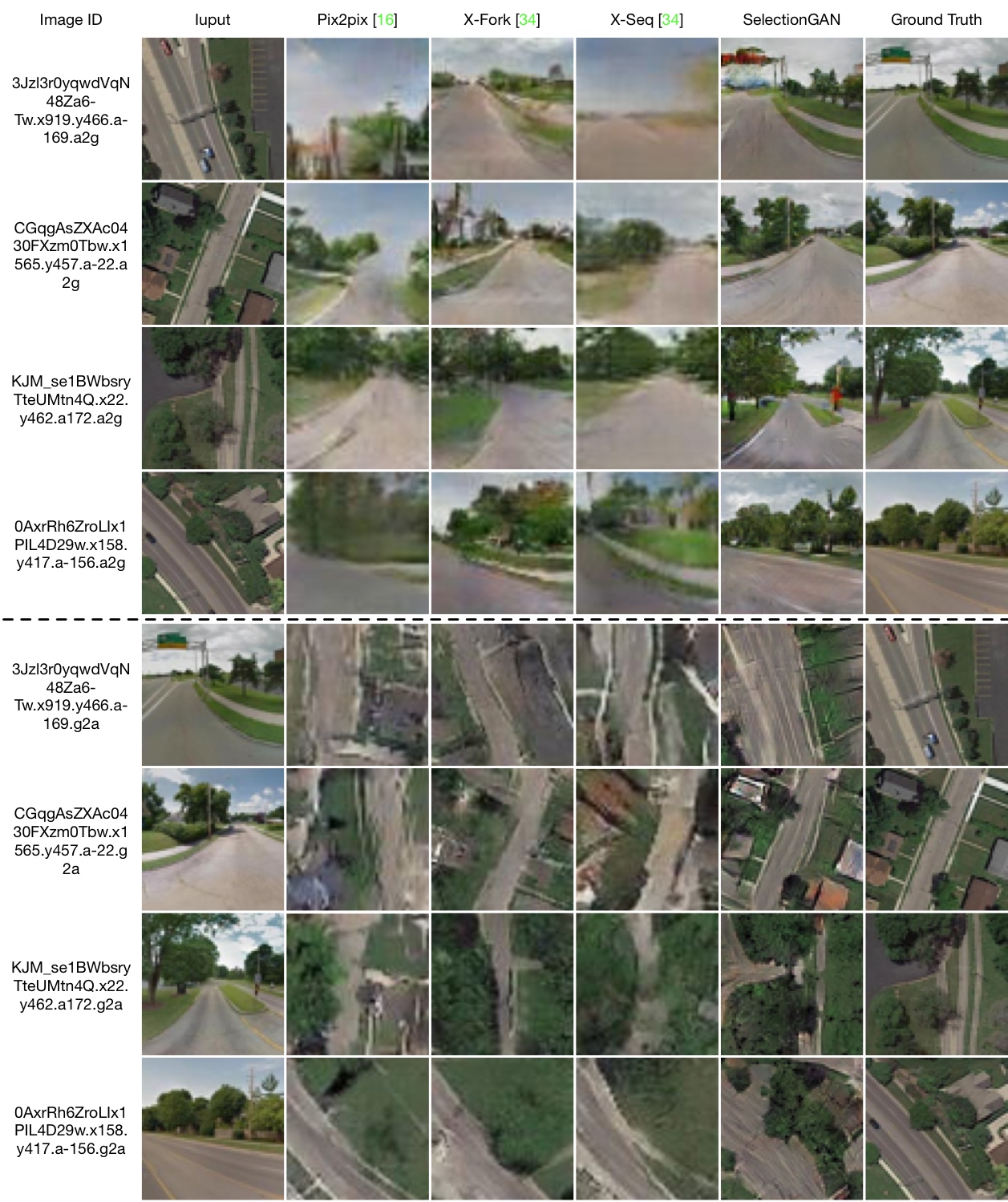}
	\caption{Results generated by different methods in $64{\times}64$ resolution in both a2g (Top) and g2a (Bottom) directions on Dayton dataset. These samples were randomly selected for visualization purposes.
	}
	\label{fig:dayton_64}
\end{figure*}

\begin{figure*}[!ht] \small
	\centering
	\includegraphics[width=1\linewidth]{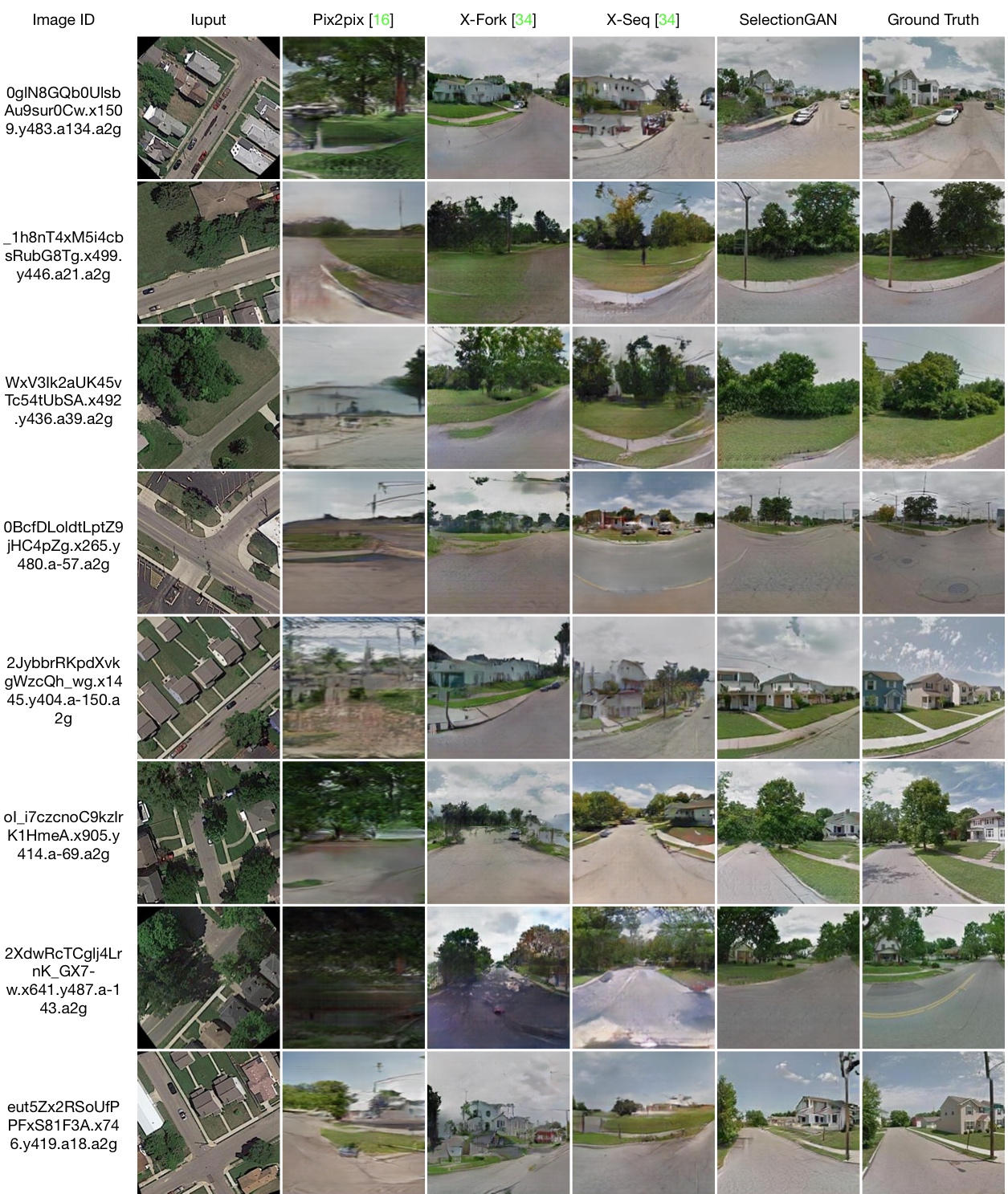}
	\caption{Results generated by different methods in $256{\times}256$ resolution in a2g direction on Dayton dataset. These samples were randomly selected for visualization purposes.
	}
	\label{fig:dayton_256_a2g}
\end{figure*}

\begin{figure*}[!ht] \small
	\centering
	\includegraphics[width=1\linewidth]{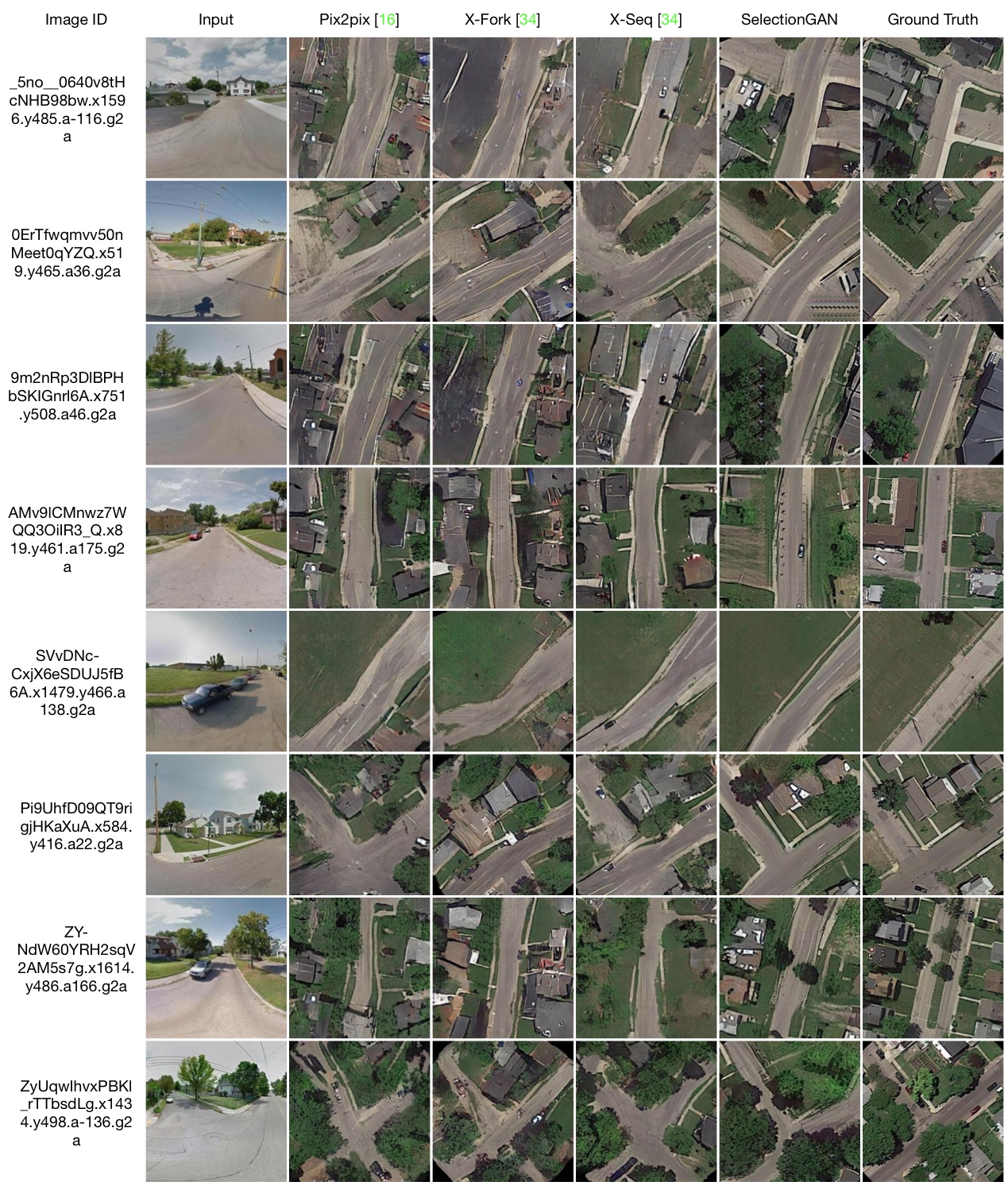}
	\caption{Results generated by different methods in $256{\times}256$ resolution in g2a direction on Dayton dataset. These samples were randomly selected for visualization purposes.
	}
	\label{fig:dayton_256_g2a}
\end{figure*}

\begin{figure*}[!ht] \small
	\centering
	\includegraphics[width=1\linewidth]{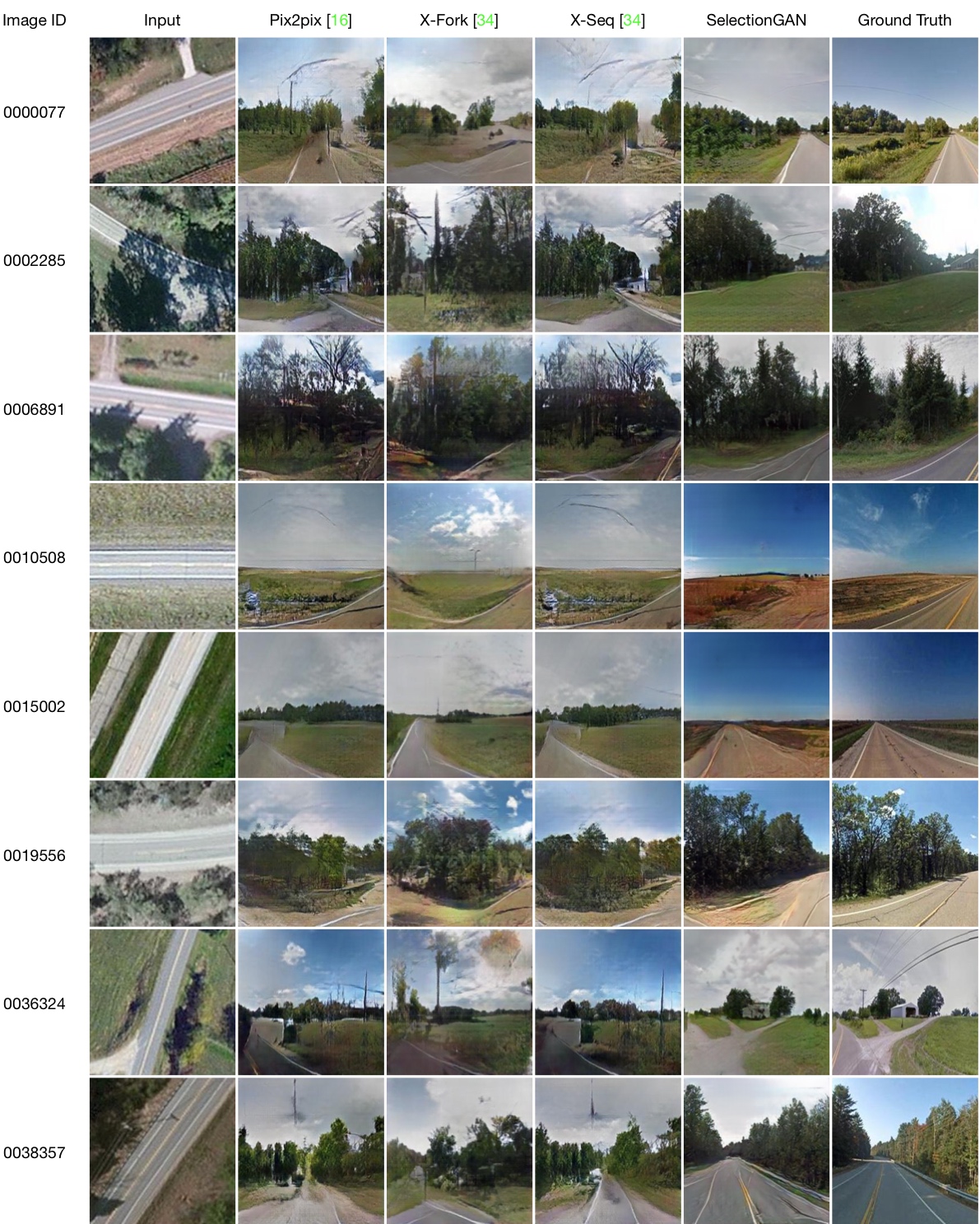}
	\caption{Results generated by different methods in $256{\times}256$ resolution in a2g direction on CVUSA dataset. These samples were randomly selected for visualization purposes.
	}
	\label{fig:cvusa_256_a2g}
\end{figure*}

\begin{figure*}[!ht] \small
	\centering
	\includegraphics[width=1\linewidth]{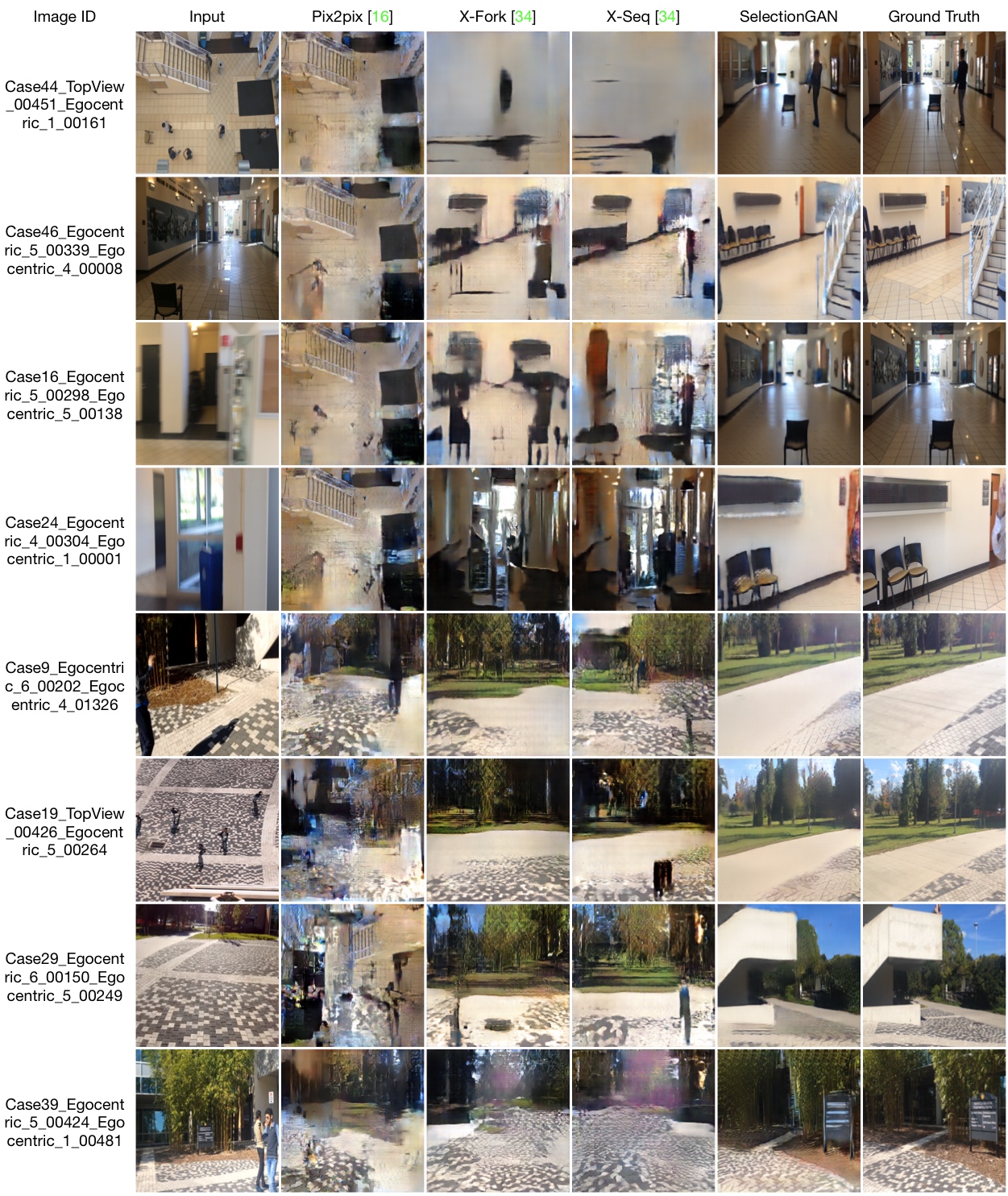}
	\caption{Results generated by different methods in $256{\times}256$ resolution on Ego2Top dataset. These samples were randomly selected for visualization purposes.
	}
	\label{fig:ego2top_256}
\end{figure*}

In Figures \ref{fig:dayton_64}, \ref{fig:dayton_256_a2g}, \ref{fig:dayton_256_g2a}, \ref{fig:cvusa_256_a2g} and \ref{fig:ego2top_256}, we show more image generation results on Dayton, CVUSA and Ego2Top datasets compared with the state-of-the-art methods \emph{i.e.}, Pix2pix~\cite{isola2017image}, X-Fork~\cite{regmi2018cross} and X-Seq~\cite{regmi2018cross}.
For Figures \ref{fig:dayton_64}, \ref{fig:dayton_256_a2g}, \ref{fig:dayton_256_g2a}, \ref{fig:cvusa_256_a2g}, we reproduced the results of Pix2pix~\cite{isola2017image}, X-Fork~\cite{regmi2018cross} and X-Seq~\cite{regmi2018cross} using the pre-trained models provided by the authors\footnote{https://github.com/kregmi/cross-view-image-synthesis}.
As we can see from all these figures, the proposed SelectionGAN achieves significantly visually better results than the competing methods.

\end{document}